\documentclass[10pt,journal,cspaper,compsoc]{IEEEtran}

\usepackage{epsfig}
\usepackage{graphicx}
\usepackage{amsmath}
\usepackage{amssymb}
\usepackage{multirow}
\usepackage{bbding}
\usepackage{pifont}
\usepackage{caption}
\usepackage[bookmarks=false,colorlinks=true,linkcolor=black,citecolor=black,filecolor=black,urlcolor=black]{hyperref}

\usepackage[subrefformat=parens]{subcaption}
\usepackage{algorithm}
\usepackage{algpseudocode}
\usepackage{amssymb}
\usepackage{bbding}
\usepackage{amsmath}
\usepackage{tabularx, booktabs} 

\newcommand{\LL}{\mathcal{L}}
\newcommand{\EE}{\mathop{\mathbb{E}}}

\begin{document}
%

\title{Adversarial-Based Knowledge Distillation for Multi-Model Ensemble and Noisy Data Refinement}

\author{Zhiqiang~Shen,~~ Zhankui~He,~~ Wanyun~Cui, ~~ Jiahui Yu,~~  \\Yutong Zheng,~~ Chenchen Zhu,~~ Marios~Savvides
\IEEEcompsocitemizethanks{
\IEEEcompsocthanksitem Zhiqiang~Shen, Yutong Zheng, Chenchen Zhu and Marios~Savvides are with the Department of Electrical and Computer Engineering, Carnegie Mellon University, Pittsburgh, PA 15213, USA. 
E-mail: \{zhiqians, yutongzh, chenchez, marioss\}@andrew.cmu.edu. 
\IEEEcompsocthanksitem Zhankui He is with the Department of Computer Science, University of California San Diego, CA 92093, USA. E-mail: zhh004@eng.ucsd.edu.
\IEEEcompsocthanksitem Wanyun Cui is with Shanghai University of Finance and Economics. E-mail: cui.wanyun@sufe.edu.cn.
\IEEEcompsocthanksitem Jiahui Yu is with the Department of Electrical and Computer Engineering, University of Illinois at Urbana-Champaign, Illinois, IL 61801, USA. 
E-mail: jyu79@illinois.edu.
}
}

\markboth{}
	{Shell \MakeLowercase{\textit{et al.}}: Bare Demo of IEEEtran.cls for Computer Society Journals}

\IEEEcompsoctitleabstractindextext{%
\begin{abstract}

Generic Image recognition is a fundamental and fairly important visual problem in computer vision. Recently with the development of modern neural network design and automatic architecture search, this task has been one of the fastest moving areas and achieved surprising results which even surpasses human-level performance on large-scale datasets like ImageNet~\cite{deng2009imagenet} and OpenImage~\cite{openimages}. 
One of the major challenges of this task lies in the fact that single image usually has multiple objects inside while the labels are still one-hot, another one is noisy and sometimes missing labels when annotated by humans.
In this paper, we focus on tackling these challenges accompanying with two different image recognition problems: multi-model ensemble and noisy data recognition with a unified framework, showing the superiority and advantages of our proposed method. As is well-known, usually the best performing deep neural models are ensembles of multiple base-level networks, as it can mitigate the variation or noise containing in the dataset. Unfortunately, the space required to store these many networks, and the time required to execute them at runtime, prohibit their use in applications where test sets are large (e.g., ImageNet).
In this paper, we present a method for compressing large, complex trained ensembles into a single network,
where the knowledge from a variety of trained deep neural networks (DNNs) is distilled and transferred to a single DNN. In order to distill diverse knowledge from different trained (teacher) models, we propose to use adversarial-based learning strategy where we define a block-wise training loss to guide and optimize the predefined student network to recover the knowledge in teacher models, and to promote the discriminator network to distinguish teacher {\em vs.} student features simultaneously. The proposed ensemble method (MEAL) of transferring distilled knowledge with adversarial learning exhibits three important advantages: (1) the student network that learns the distilled knowledge with discriminators is superiorly optimized than the original model; (2) fast inference is realized by a single forward pass, while the performance is even better than traditional ensembles from multi-original models; (3) the soft distributions from teacher networks can provide more informative and accurate supervision signals in the image for student learning, which can overcome the multi-object and noisy label problem to some extent. 
Extensive experiments on CIFAR-10/100, SVHN, ImageNet and iMaterialist Challenge Dataset demonstrate the effectiveness of our MEAL method. On ImageNet, our ResNet-50 based MEAL achieves top-1/5 21.79\%/5.99\% val error, which outperforms the original model by 2.06\%/1.14\%. On iMaterialist Challenge Dataset, our MEAL obtains a remarkable improvement of top-3 1.15\% (official evaluation metric) on a strong baseline model of ResNet-101. 

\end{abstract}

}

\newcommand{\methodname}{dense convolutional  network}
\newcommand{\methodnamecap}{Dense Convolutional Network}
\newcommand{\methodnameshort}{DenseNet}
\newcommand{\methodnameshorts}{DenseNets}
\newcommand{\methodblock}{dense block}
\newcommand{\methodblockcap}{Dense Block}

\newcommand{\regmethodname}{feature drop}
\newcommand{\regmethodnamecap}{Feature Drop}

\newcommand{\stepsizename}{growth rate}

\newcommand{\conv}[1]{$\left[\begin{array}{ll} \text{1}\times \text{1} \text{ conv}\\ \text{3}\times \text{3} \text{ conv} \end{array}\right] \times \text{#1}$}

\newcommand{\cross}[1]{#1 $\times$ #1}

\newcommand{\feati}{x_i}
\newcommand{\clsfeati}{y_i}
\newcommand{\featk}{x_k}
\newcommand{\clsfeatk}{y_k}
\newcommand{\loss}{L}
\newcommand{\featL}{x_L}
\newcommand{\clsfeat}{y}
\newcommand{\anyxs}{\ensuremath{\mathbf{x}}}
\newcommand{\anyys}{\ensuremath{\mathbf{y}}}

\newcommand{\bx}{\ensuremath{\mathbf{x}}}

\newcommand{\sourcexs}{\ensuremath{\mathbf{x^\mathcal{S}}}}
\newcommand{\sourceys}{\ensuremath{\mathbf{y^\mathcal{S}}}}

\newcommand{\targetxs}{\ensuremath{\mathbf{x^\mathcal{T}}}}
\newcommand{\targetys}{\ensuremath{\mathbf{y^\mathcal{T}}}}
\newcommand{\pseudotargetys}{\ensuremath{\mathbf{\hat{y}^\mathcal{T}}}}

\maketitle


\IEEEpeerreviewmaketitle

\section{Introduction}\label{sec:introduction}

\IEEEPARstart{T}{he} model ensemble approach is a collection of neural networks whose predictions are combined at test stage by weighted averaging or voting.
It has been long observed that ensembles of multiple networks are generally much more robust and accurate than a single network if the training data is noisy and intractable to handle. This benefit has also been exploited indirectly when training a single network through Dropout~\cite{srivastava2014dropout}, Dropconnect~\cite{wan2013regularization}, Stochastic Depth~\cite{huang2016deep}, Swapout~\cite{singh2016swapout}, etc. We extend this idea by forming ensemble predictions during training, using the outputs of different network architectures with different or identical augmented input. Our testing still operates on a single network, but the supervision labels made on different pre-trained networks correspond to an ensemble prediction of a group of individual reference networks.

The traditional ensemble, or called true ensemble, has some disadvantages that are often overlooked. 1) Redundancy: The information or knowledge contained in the trained neural networks are always redundant and has overlaps between with each other. Directly combining the predictions often requires extra computational cost but the gain is limited. 2) Ensemble is always large and slow: Ensemble requires more computing operations than an individual network, which makes it unusable for applications with limited memory, storage space, or computational power such as  desktop, mobile and even embedded devices, and for applications in which real-time predictions are needed.

\begin{figure}[t]
	\centering
	\includegraphics[width=0.39\textwidth]{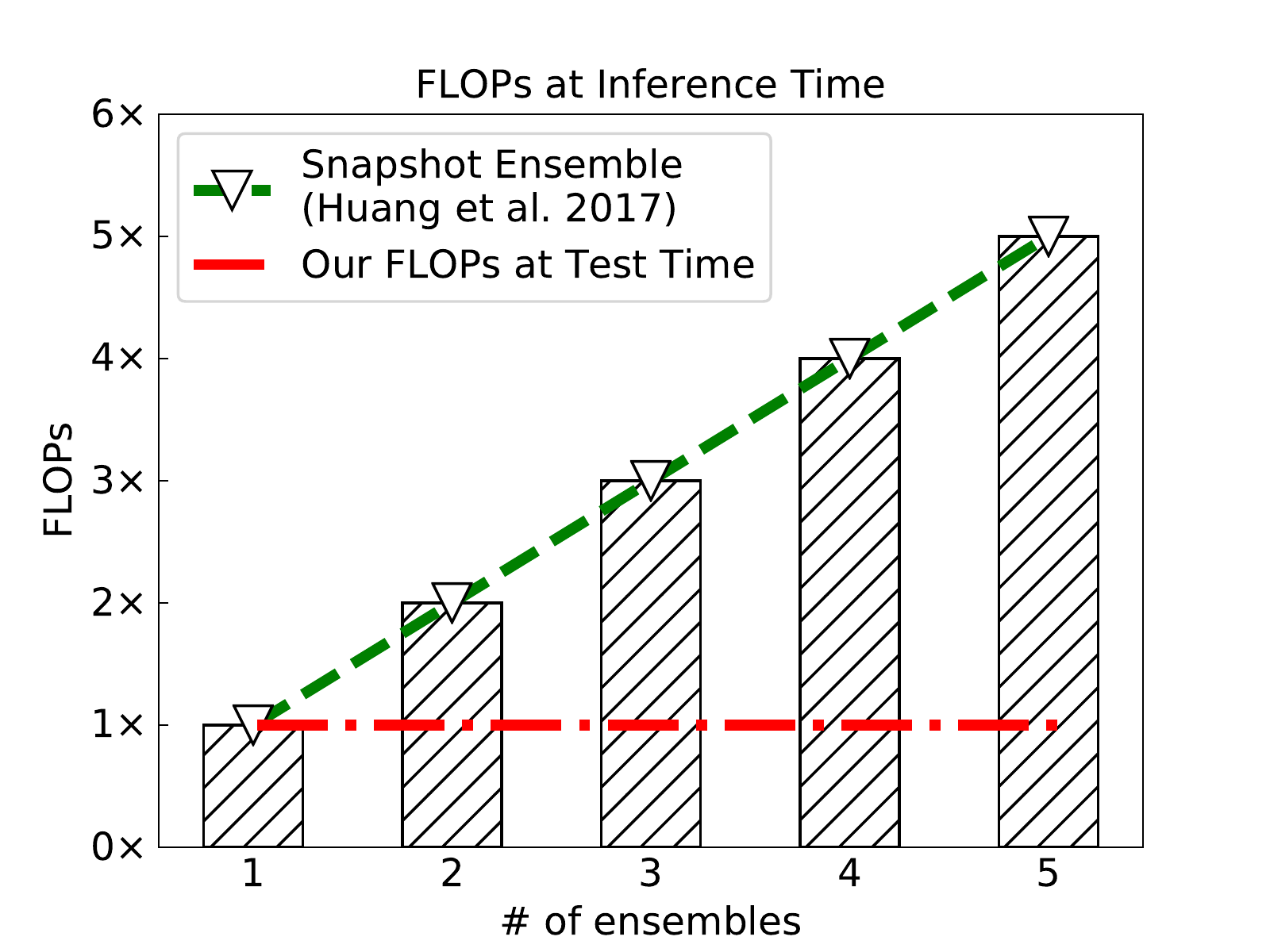}
	\vspace{-0.10in}
	\caption{Comparison of FLOPs at inference time. Huang et al.~\cite{huang2017snapshot} employ models at different local minimum for ensembling, which enables no additional training cost, but the computational FLOPs at test time linearly increase with more ensembles. In contrast, our method use only one model during inference time throughout, so the testing cost is independent of \# ensembles.}
	\label{inference_time}
	\vspace{-0.1in}
\end{figure}

\begin{figure}[t]
	\centering
	\includegraphics[width=0.49\textwidth]{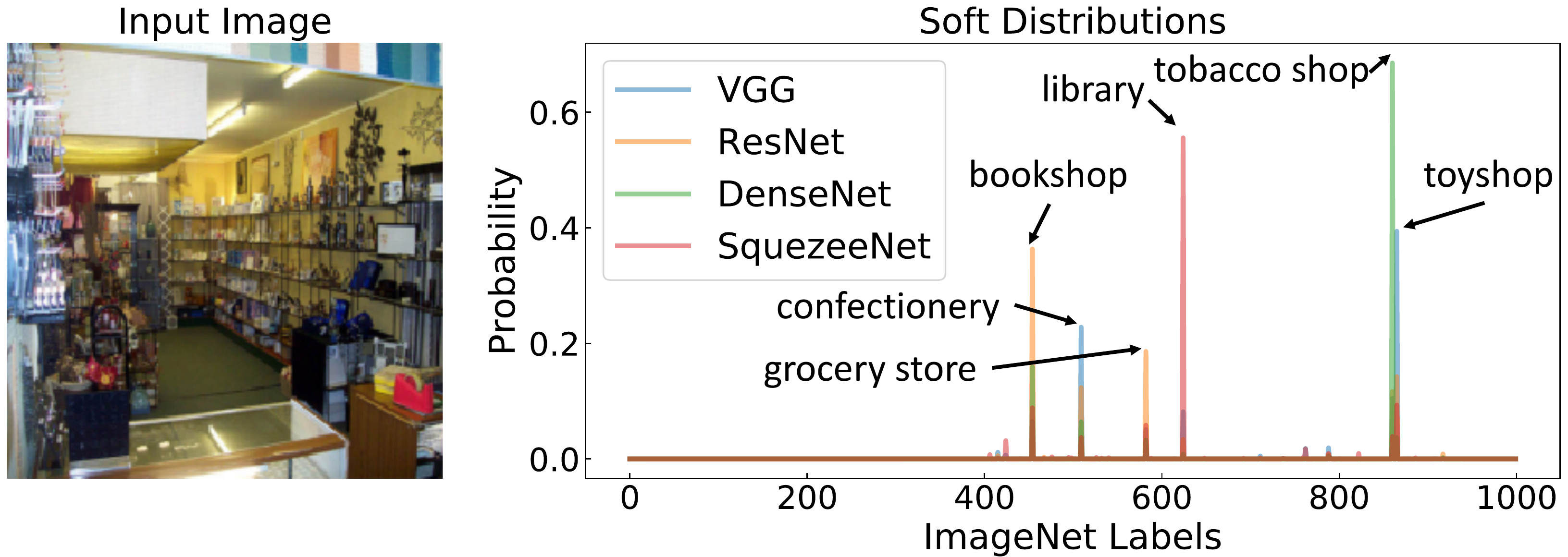}
	\caption{Left is a training example of class ``tobacco shop'' from ImageNet. Right are soft distributions from different trained architectures. The soft labels are more {\em {informative}} and can provide more coverage for visually-related scenes.}
	\label{soft_labels}
\end{figure}

To address the aforementioned shortcomings, in this paper we propose to use a learning-based ensemble method. Our goal is to learn an ensemble of multiple neural networks without incurring any additional {\em {testing costs}}, as shown in Fig.~\ref{inference_time}. We achieve this goal by leveraging the combination of diverse outputs from different neural networks as supervisions to guide the target network training. The reference networks are called {\em {Teachers}} and the target networks are called {\em {Students}}. Instead of using the traditional one-hot vector labels, we use the {\em {soft}} labels that provide more coverage for co-occurring and visually related objects and scenes. We argue that labels should be informative for the specific image. In other words, the labels should not be identical for all the given images with the same class. More specifically, as shown in Fig.~\ref{soft_labels}, an image of ``tobacco shop'' has similar appearance to ``library'' should have a different label distribution than an image of ``tobacco shop'' but is more similar to ``grocery store''. It can also be observed that soft labels can provide the additional intra- and inter-category relations of datasets.

\begin{figure}[t]
	\begin{subfigure}[b]{.49\linewidth}
	\centering
	\includegraphics[width=0.98\textwidth]{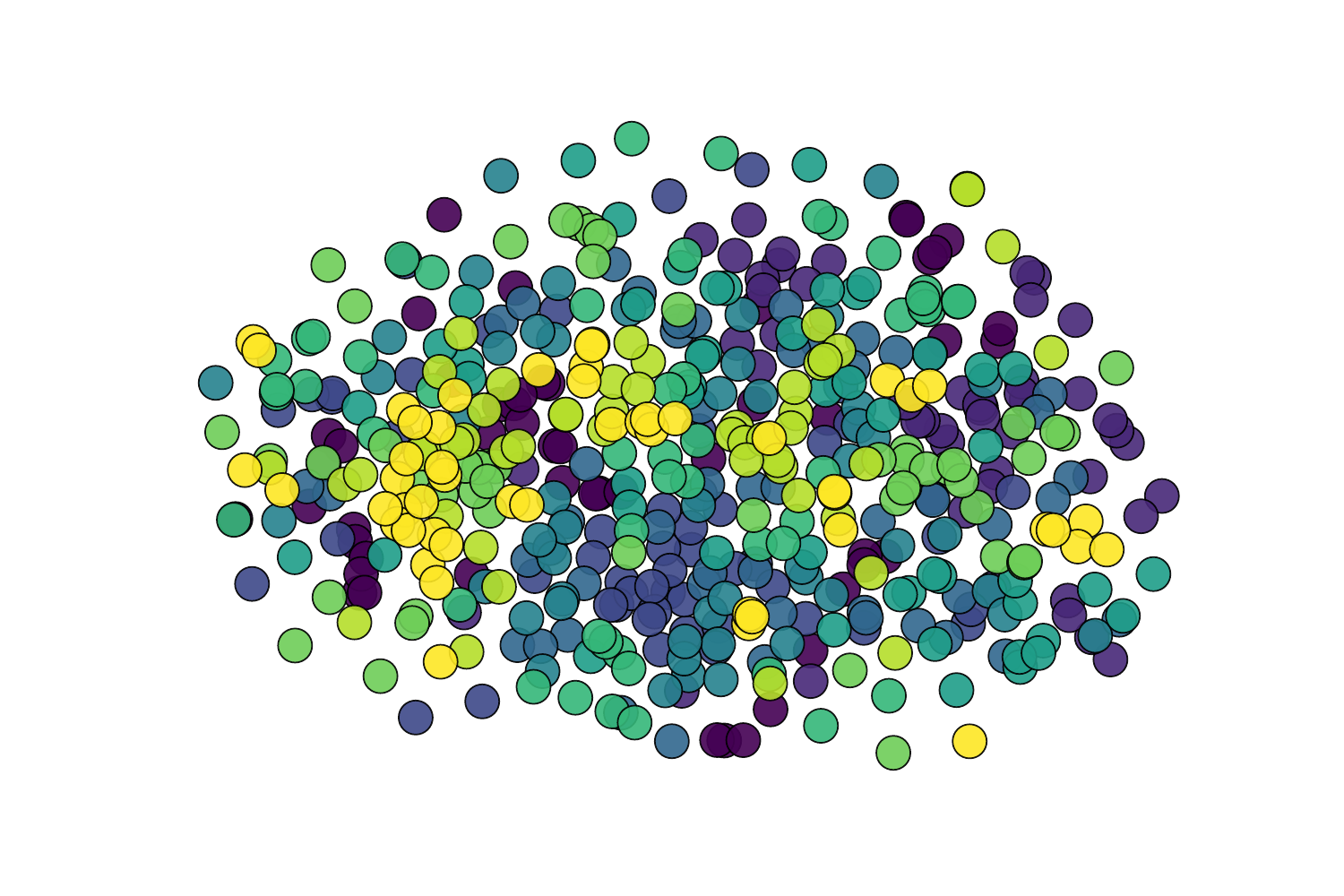}
	\caption{Standard}\label{fig1a}
	\end{subfigure}\hfill
	\begin{subfigure}[b]{.49\linewidth}
	\includegraphics[width=0.98\textwidth]{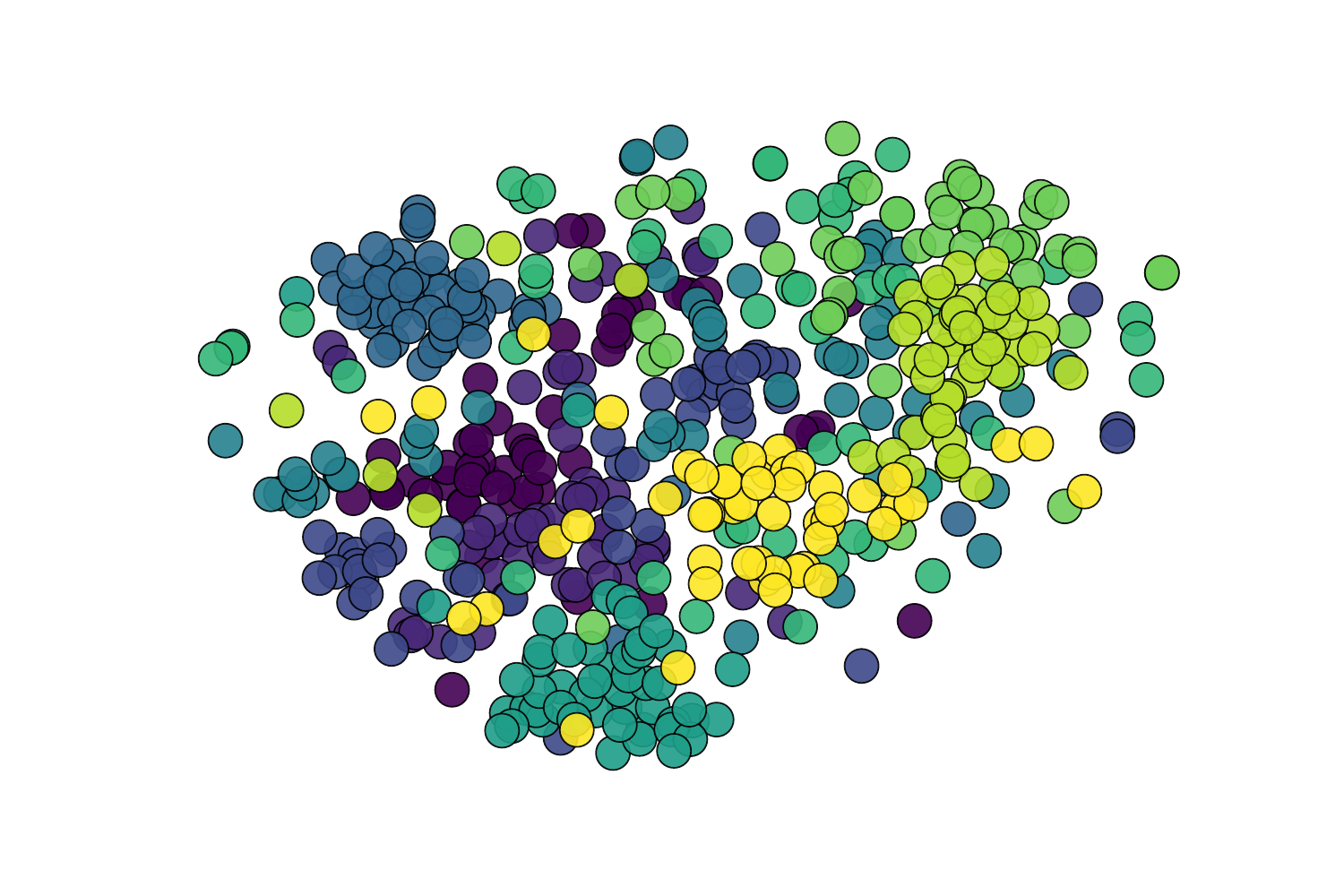}
	\caption{Ours}\label{fig1a}
    \end{subfigure}\hfill
	\caption{Visualizations of validation images from the ImageNet dataset~\cite{deng2009imagenet} by t-SNE~\cite{maaten2008visualizing}. We randomly sample 10 classes within 1000 classes. Left is the single model result using the standard training strategy. Right is our MEAL ensemble model result.}
	\label{embedding}
\end{figure}

To further improve the robustness of student networks, we introduce an adversarial learning strategy to force the student to generate similar outputs as teachers. We propose two different strategies for the generative adversarial training: (i) joint training with a unified framework; and (ii) alternately update gradients with separate training processes, i.e., updating the gradients in discriminator and student network iteratively. To the best of our knowledge, there are very few existing works adopting generative adversarial learning to force the student networks  to have similar distribution outputs with the teachers, so our proposed method is a pioneer of this direction for multi-model ensemble.
Our experiments show that MEAL consistently improves the accuracy across a variety of popular network architectures on different datasets. For instance, our shake-shake~\cite{gastaldi2017shake} based MEAL achieves 2.54\% test error on CIFAR-10, which is a relative $11.2\%$ improvement\footnote{Shake-shake baseline~\cite{gastaldi2017shake} is 2.86\%.}. On ImageNet, our ResNet-50 based MEAL achieves 21.79\%/5.99\% val error, which outperforms the baseline by a large margin.

Furthermore, we extend our method to the problem of noisy data processing. We propose an iterative refinement paradigm based on our MEAL method, which can refine the labels from the teacher networks progressively and provide more accurate supervisions for the student network training. We conduct experiments on iMaterialist Challenge Dataset and the results show that our method can vastly improve the performance of base models.

To explore what our model actually learned, we visualize the embedded features from the single model and our ensembling model. The visualization is plotted by t-SNE tool~\cite{maaten2008visualizing} with the last conv-layer features (2048 dimensions) from ResNet-50. We randomly sample 10 classes on ImageNet, results are shown in Fig.~\ref{embedding}, it's obvious that our model has better feature embedding result.

In summary, our contribution in this paper is three fold.

\begin{itemize}
\addtolength{\itemsep}{-0.0in}
\item  An end-to-end framework with adversarial learning is designed based on the {\em {teacher-student}} learning paradigm for deep neural network ensembling and noisy data learning.
\item The proposed method can achieve the goal of ensembling multiple neural networks with no additional {\em {testing cost}}.
\item The proposed method improves the state-of-the-art accuracy on CIFAR-10/100, SVHN, ImageNet and iMaterialist Challenge Dataset for a variety of existing network architectures.
\end{itemize}

A preliminary version of this manuscript~\cite{shen2019MEAL} has been published in a previous conference. In this version, we involved and compared two different gradient update strategies for adversarial learning on our proposed MEAL framework. We also provided a novel learning paradigm for how to adopt our method on handling noisy date circumstances. Furthermore, we included more experiments, details, analysis and an iterative refinement strategy with better performance. Currently, there are few works focusing on adopting generative adversarial learning on feature space for learning identical distributions between teacher and student networks. Thus, this work gives very good and practical guidelines for multi-model learning/ensemble and noisy data refinement.

\section{Related Work}\label{related_work}
There is a large body of previous work~\cite{hansen1990neural,perrone1995networks,krogh1995neural,dietterich2000ensemble,huang2017snapshot,lakshminarayanan2017simple,zhu2018knowledge,zhou2018diverse} on ensembles with neural networks. However, most of these prior studies focus on improving the generalization of an individual network. Recently, Snapshot Ensembles~\cite{huang2017snapshot} is proposed to address the cost of training ensembles. In contrast to the Snapshot Ensembles, here we focus on the cost of {\em {testing ensembles}}. Our method is based on the recently raised knowledge distillation~\cite{hinton2015distilling,papernot2016semi,li2017learning,yim2017gift} and adversarial learning~\cite{goodfellow2014generative}, so we will review the ones that are most directly connected to our work.

\vspace{.5em}
\noindent{\textbf{``Implicit'' Ensembling.}}
Essentially, our method is an ``implicit'' ensemble which usually has high efficiency during both training and testing. The typical ``implicit'' ensemble methods include: Dropout~\cite{srivastava2014dropout}, DropConnection~\cite{wan2013regularization}, Stochastic Depth~\cite{huang2016deep}, Swapout~\cite{singh2016swapout}, etc. These methods generally create an exponential number of networks with shared weights during training and then implicitly ensemble them at test time. In contrast, our method focuses on the subtle differences of labels with identical input. Perhaps the most similar to our work is the recent proposed Label Refinery~\cite{bagherinezhad2018label}, who focus on the single model refinement using the softened labels from the previous trained neural networks and iteratively learn a new and more accurate network. Our method differs from it in that we introduce adversarial modules to force the model to learn the difference between teachers and students, which can improve model generalization and can be used in conjunction with any other implicit ensembling techniques. There are some other ensemble methods like DivE$^2$~\cite{zhou2018diverse}, which aims to train an ensemble of models that assigns data to models at each training epoch based on each model’s current expertise and an intra- and inter-model diversity reward. It starts by choosing easy samples for each
model, and then gradually adjusts towards the models having specialized and complementary expertise on subsets of the training data. 

\vspace{.5em}
\noindent{\textbf{Adversarial Learning.}}
Generative Adversarial Learning~\cite{goodfellow2014generative} is firstly proposed to generate realistic-looking images from random noise using neural networks. It consists of two components. One serves as a generator and another one as a discriminator. The generator is used to synthesize images to fool the discriminator, meanwhile, the discriminator tries to distinguish real and fake images. Recently, numerous interesting GAN evolution algorithms have been proposed, such as Wasserstein GAN~\cite{arjovsky2017wasserstein}, Improved Wasserstein gans~\cite{gulrajani2017improved}, DRAGAN~\cite{kodali2017convergence}, NS GAN~\cite{fedus2017many}, LS GAN~\cite{mao2017least}, et al. Generally, the generator and discriminator are trained simultaneously through competing with each other. In this work, we employ generators to synthesize student features and use discriminator to discriminate between teacher and student outputs for the same input image. An advantage of adversarial learning is that the generator tries to produce similar features as a teacher that the discriminator cannot differentiate. This procedure improves the robustness of training for student network and has applied to many fields such as image-to-image translation~\cite{isola2017image,zhu2017unpaired,zhu2017toward,liu2017unsupervised,huang2018multimodal,shen2019towards}, image generation~\cite{johnson2018image}, detection~\cite{bai2018finding}, etc.

\begin{figure*}[t]
	\centering
	\includegraphics[width=0.80\textwidth]{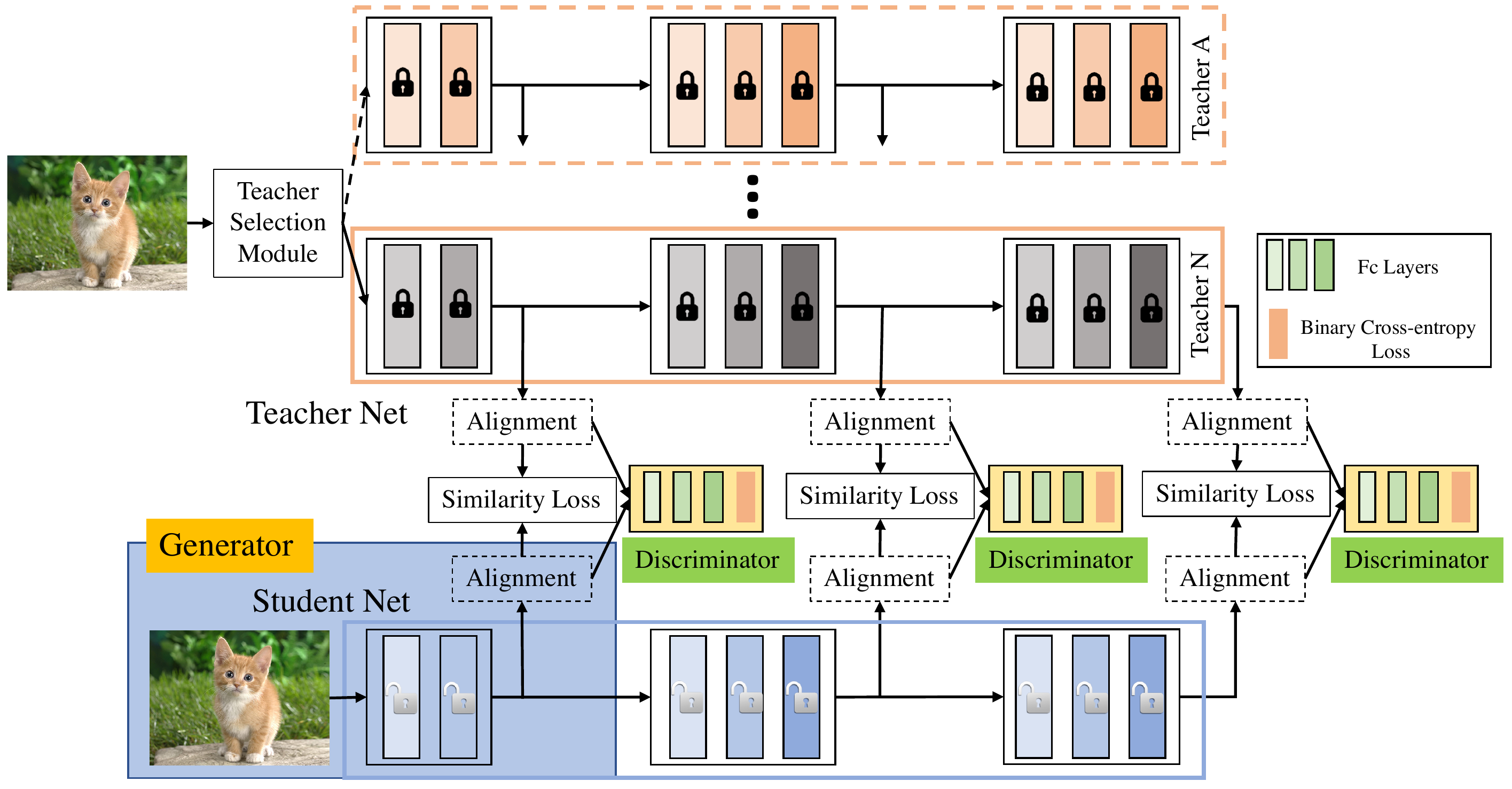}
	\caption{Overview of our proposed architecture. We input the same image into the teacher and student networks to generate intermediate and final outputs for {\em {Similarity Loss}} and {\em {Discriminators}}. The model is trained adversarially against several {\em {discriminator networks}}. During training the model observes supervisions from trained teacher networks instead of the one-hot ground-truth labels, and the teacher's parameters are fixed all the time.}
	\label{al}
\end{figure*}

\vspace{.5em}
\noindent{\textbf{Knowledge Transfer.}}
Distilling knowledge from trained neural networks and transferring it to another new network has been well explored in~\cite{hinton2015distilling,chen2015net2net,li2017learning,xu2017training,yim2017gift,tarvainen2017mean,bagherinezhad2018label,anil2018large,chen2018darkrank,park2019relational,heo2018knowledge,furlanello2018born,uijlings2018revisiting,crowley2018moonshine,yang2019snapshot}. The typical way of transferring knowledge is the {\em {teacher-student learning paradigm}}, which uses a softened distribution of the final output of a teacher network to teach information to a student network. With this teaching procedure, the student can learn how a teacher studied given tasks in a more efficient form. Yim et al.~\cite{yim2017gift} defined the distilled knowledge to be transferred flows between different intermediate layers and computered the inner product between parameters from two networks. Bagherinezhad et al.~\cite{bagherinezhad2018label} studied the effects of various properties of labels and introduce the {\em {Label Refinery}} method that iteratively updated the ground truth labels after examining the entire dataset with the {{teacher-student learning paradigm}}. Park et al.~\cite{park2019relational} introduced a novel dubbed relational knowledge distillation (RKD) that transferred mutual relations of data examples instead. For concrete realizations of RKD, they proposed distance-wise and angle-wise distillation losses that penalize structural differences in relations.

\vspace{.5em}
\noindent{\textbf{Learning with Noisy Labels.}}   Learning with noisy labels has been a widely-explored research topic in the recent years, since it has wide usage and applications~\cite{li2019learning}. There are a large number of variety methods which are applied to tackle this problem of modeling distribution of noisy and true annotations, such as knowledge graphs and distillation~\cite{li2017learning}, conditonal random fields~\cite{vahdat2017toward}, directed graphical models~\cite{xiao2015learning}, etc. As in the deep learning era, one line to handle this problem with neural networks is ~\cite{li2019learning,jiang2017mentornet,goldberger2016training,veit2017learning,sukhbaatar2014training,patrini2017making,ren2018learning,lee2018cleannet}, which are formulated by explicit or implicit noisy models. 

In particular, Li et al.~\cite{li2019learning} proposed to use a unified distillation framework to adopt “side” information, including a small clean dataset and label relations in knowledge graph, to “hedge the risk” of learning from noisy labels. Liu et al.~\cite{liu2015classification}  presented an importance reweighting framework for classification in the presence of label noise which sample labels are randomly corrupted. Li et al.~\cite{li2019learning} proposed a noise-tolerant learning algorithm, which a meta-learning update strategy is performed prior to conventional gradient update operation. The proposed meta-learning method simulates actual training by generating synthetic noisy labels, and train the model after one gradient update using each set of synthetic noisy labels. Sukhbaatar et al.~\cite{sukhbaatar2014training} explored the performance of discriminatively-trained CNN when training on noisy data. They introduced an extra noise layer into the network which adapted the network outputs to match the noisy label distribution. The parameters of this noise layer can be estimated as part of the training process and involve simple modifications to current training infrastructures for deep neural networks.

\section{Overview}
\noindent{\textbf{Siamese-like Network Structure}}
Our framework is a siamese-like architecture that contains two-stream networks in teacher and student branches. The structures of two streams can be identical or different, but should have the same number of blocks, in order to utilize the intermediate outputs. 
The whole framework of our method is shown in Fig.~\ref{al}. It consists of a teacher network, a student network, alignment layers, similarity loss layers and discriminators.

The teacher and student networks are processed to generate intermediate outputs for alignment. The alignment layer is an adaptive pooling process that takes the same or different length feature vectors as input and output fixed-length new features. We force the model to output similar features of student and teacher by training student network adversarially against several discriminators.
We will elaborate each of these components in the following sections with more details.
\section{Adversarial Learning (AL) for Knowledge Distillation}

\subsection{Similarity Measurement}
Given a dataset $\mathcal{D}={(X_i, Y_i)}$, we pre-trained the teacher network $\mathcal{T}_\theta$ over the dataset using the cross-entropy loss against the one-hot image-level labels{\footnote{Ground-truth labels}} in advance. The student network $\mathcal{S}_\theta$ is trained over the same set of images, but uses labels generated by $\mathcal{T}_\theta$. More formally, we can view this procedure as training $\mathcal{S}_\theta$ on a new labeled dataset $\tilde{\mathcal{D}}={(X_i, \mathcal{T}_\theta(X_i))}$. Once the teacher network is trained, we freeze its parameters when training the student network. 

We train the student network $\mathcal{S}_\theta$ by minimizing the similarity distance between its output and the soft label generated by the teacher network. Letting $p_c^{{\mathcal T}_\theta}({X_i})={\mathcal{T}_\theta }({X_i})[c]$, $p_c^{{\mathcal S}_\theta}({X_i})={\mathcal{S}_\theta }({X_i})[c]$ be the probabilities assigned to class $c$ in the teacher model $\mathcal T_\theta$ and student model $\mathcal S_\theta$. The similarity metric can be formulated as:
\begin{equation}
	\begin{gathered}
		\LL_{Sim} = d({\mathcal{T}_\theta }({X_i}),{\mathcal{S}_\theta }({X_i})) \hfill \\
		\quad \quad \;\; = \sum\limits_c {d(p_c^{\mathcal{T}_\theta}({X_i}),p_c^{\mathcal{S}_\theta}({X_i}))}  \hfill \\ 
	\end{gathered} 
\end{equation}
We investigated three distance metrics in this work, including $\ell_1$, $\ell_2$ and KL-divergence. The detailed experimental comparisons are shown in Tab.~\ref{ablation}. Here we formulate them as follows.

\noindent{\textbf{$\ell_1$ distance}} is used to minimize the absolute differences between the estimated student probability values and the reference teacher probability values. Here we formulate it as:
\begin{equation}
	\LL_{\ell_1\_Sim}({\mathcal{S}_\theta }) = \frac{1}{n}\sum\limits_c {\sum\limits_{i = 1}^n {{{\left| {p_c^{\mathcal{T}_\theta }({X_i}) - p_c^{\mathcal{S}_\theta }({X_i})} \right|}^1}} } 
\end{equation}

\noindent{\textbf{$\ell_2$ distance}} or euclidean distance is the straight-line distance in euclidean space, which has been used in Mean Teacher~\cite{tarvainen2017mean} (mean squared error, MSE) as the consistency loss. We use $\ell_2$ loss function to minimize the error which is the sum of all squared differences between the student output probabilities and the teacher probabilities. The $\ell_2$ can be formulated as:
\begin{equation}
	\LL_{\ell_2\_Sim}({\mathcal{S}_\theta }) = \frac{1}{n}\sum\limits_c {\sum\limits_{i = 1}^n {{{\left\| {p_c^{\mathcal{T}_\theta }({X_i}) - p_c^{\mathcal{S}_\theta }({X_i})} \right\|}^2}}} 
\end{equation}

\noindent{\textbf{KL-divergence}} is a measure of how one probability distribution is different from another reference probability distribution. Here we train student network $\mathcal{S}_\theta$ by minimizing the KL-divergence between its output $p_c^{{\mathcal S}_\theta}({X_i})$ and the soft labels $p_c^{{\mathcal T}_\theta}({X_i})$ generated by the teacher network. Our loss function is:
\begin{equation}
	\begin{gathered}
		{\LL_{KL\_Sim}}({\mathcal{S}_\theta }) =  - \frac{1}{n}\sum\limits_c {\sum\limits_{i = 1}^n {p_c^{{\mathcal{T}_\theta }}({X_i})\log (\frac{{p_c^{{S_\theta }}({X_i})}}{{p_c^{{\mathcal{T}_\theta }}({X_i})}})} }  \hfill \\
		\quad \quad \quad  \quad \quad \quad    =  - \frac{1}{n}\sum\limits_c {\sum\limits_{i = 1}^n {p_c^{{\mathcal{T}_\theta }}({X_i})\log } } p_c^{{\mathcal{S}_\theta }}({X_i}) \hfill \\
		\quad \quad \quad  \quad \quad \quad \quad  + \frac{1}{n}\sum\limits_c {\sum\limits_{i = 1}^n {p_c^{{\mathcal{T}_\theta }}({X_i})\log } } p_c^{{\mathcal{T}_\theta }}({X_i}) \hfill \\ 
	\end{gathered} 
\end{equation}
where the second term is the entropy of soft labels from teacher network and is constant with respect to $\mathcal{T}_\theta$. We can remove it and simply minimize the cross-entropy loss as follows:
\begin{equation}
	{{{\LL}}_{CE\_Sim}}({\mathcal{S}_\theta }) =  - \frac{1}{n}\sum\limits_c {\sum\limits_{i = 1}^n {p_c^{{\mathcal{T}_\theta }}({X_i})\log } } p_c^{{\mathcal{S}_\theta }}({X_i})
\end{equation}

\subsection{Intermediate Alignment}
\noindent{\textbf{Adaptive Pooling.}} The purpose of the adaptive pooling layer is to align the intermediate output from teacher network and student network. This kind of layer is similar to the ordinary pooling layer like average or max pooling, but can generate a predefined length of output with different input size. Because of this specialty, we can use the different teacher networks and pool the output to the same length of student output. Pooling layer can also achieve spatial invariance when reducing the resolution of feature maps. Thus, for the intermediate output, our loss function is:
\begin{equation}
	\LL^j_{Sim} = d(f(\mathcal{T}_{{\theta}_j}), f(\mathcal{S}_{{\theta}_j}))
\end{equation}
where $T_{{\theta}_j}$ and $S_{{\theta}_j}$ are the outputs at $j$-th layer of the teacher and student, respectively. $f$ is the adaptive pooling function that can be average or max. Fig.~\ref{ap} illustrates the process of  adaptive pooling. Because we adopt multiple intermediate layers, our final similarity loss is a sum of individual one:
\begin{equation} \label{sim}
	\LL_{Sim}  = \sum\limits_{j\in {\mathcal{A}} } {\LL^j_{Sim}}
\end{equation}
where $\mathcal{A}$ is the set of layers that we choose to produce output. In our experiments, we use the last layer in each block of a network (block-wise).

\begin{figure}[t]
	\centering
	\includegraphics[width=0.48\textwidth]{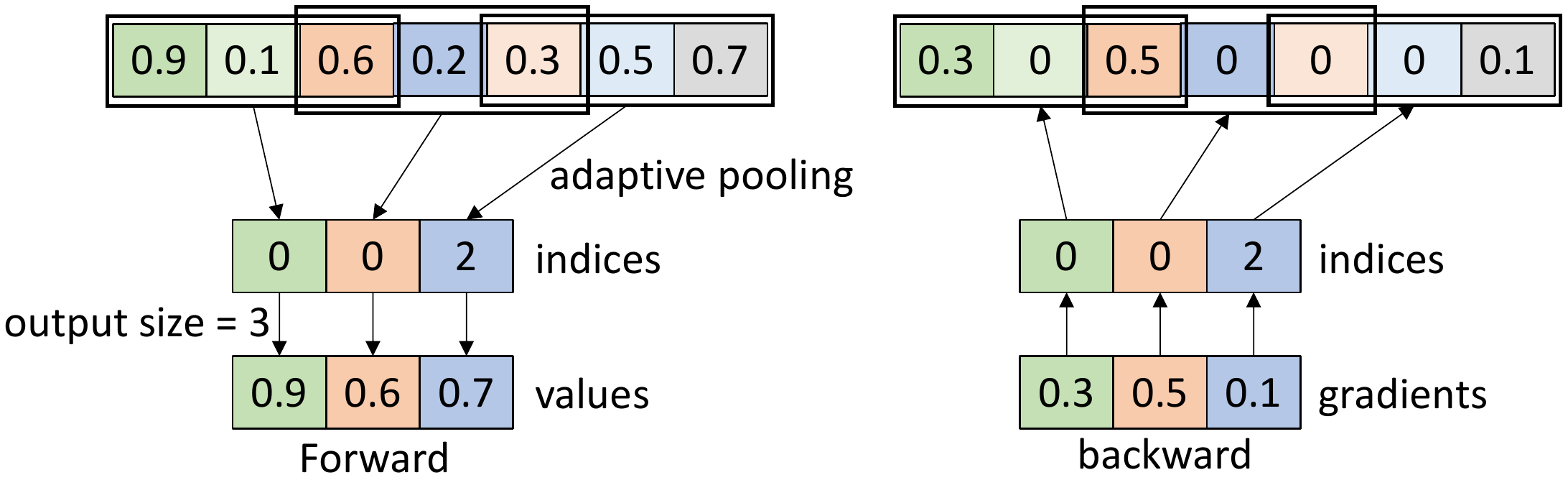}
	\caption{The process of adaptive pooling in forward and backward stages. We use max operation for illustration.}
	\label{ap}
\end{figure}

\subsection{Stacked Discriminators} 

We generate student output by training the student network $\mathcal{S}_\theta$ and freezing the teacher parts adversarially against a series of stacked discriminators {\em D}$_j$. A discriminator {\em D} attempts to classify its input $x$ as teacher or student by maximizing the following objective as in~\cite{goodfellow2014generative}:
\begin{equation}
	\LL^j_{GAN} = \EE_{x\sim p_{\textrm{teacher}}} \log D_j(x) + \EE_{x\sim p_{\textrm{student}}} \log(1 - D_j(x))
\end{equation}
where ${x\sim p_{\textrm{student}}}$ are outputs from generation network $\mathcal{S}_{{\theta}_j}$. At the same time, $\mathcal{S}_{{\theta}_j}$ attempts to generate similar outputs which will fool the discriminator by minimizing $\EE_{x\sim p_{\textrm{student}}} \log(1 - D_j(x))$.

\begin{figure}[t]
	\centering
	\includegraphics[width=0.48\textwidth]{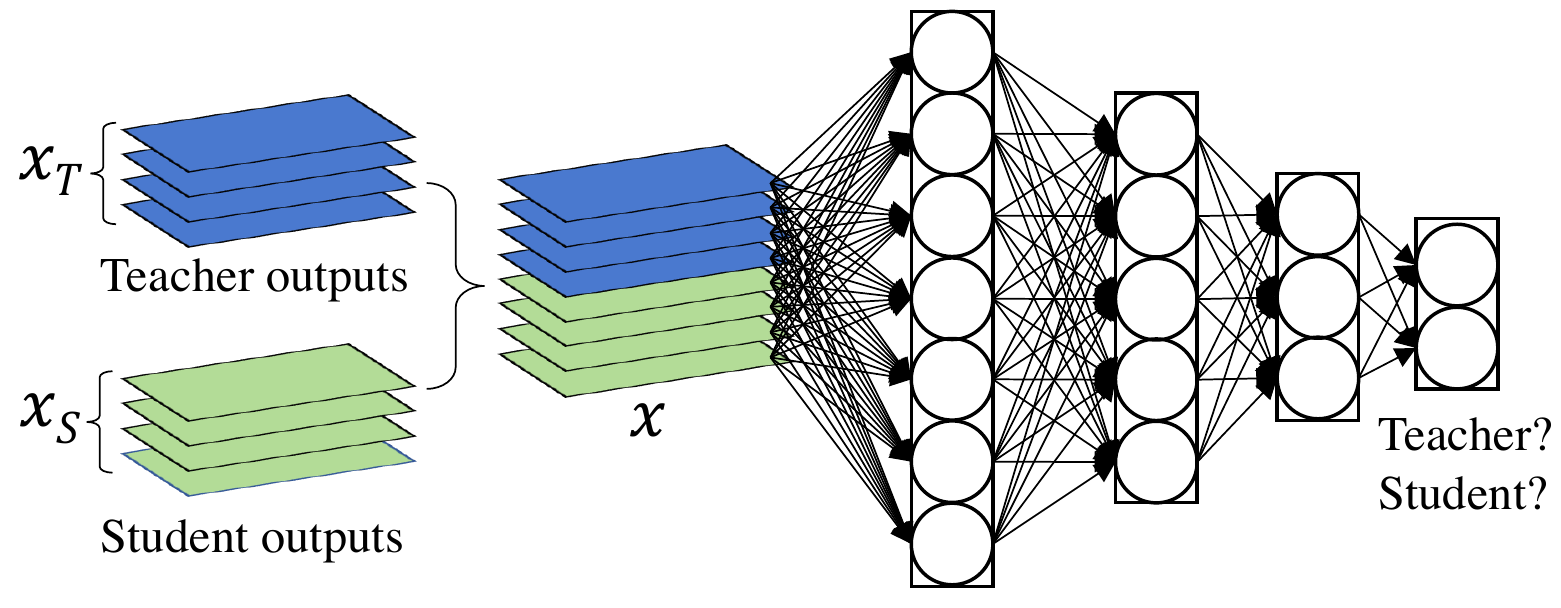}
	\caption{Illustration of our proposed discriminator. We concatenate the outputs of teacher and student as the inputs of a discriminator. The discriminator is a three-layer fully-connected network.}
	\label{d}
\end{figure}

In Eq.~\ref{gan}, $x$ is the concatenation of teacher and student outputs. We feed $x$ into the discriminator which is a three-layer fully-connected network. The whole structure of a discriminator is shown in Fig.~\ref{d}.

\noindent{\textbf{Multi-Stage Discriminators.}}
Using multi-Stage discriminators can refine the student outputs gradually.
As shown in Fig.~\ref{al}, the final adversarial loss is a sum of the individual ones (by minimizing -$\LL^j_{GAN}$):
\begin{equation} \label{gan}
	\LL_{GAN}  = -\sum\limits_{j\in {\mathcal{A}}} { \LL^j_{ GAN}}
\end{equation}
Let $\left| \mathcal{A} \right|$ be the number of discriminators. In our experiments, we use 3 for CIFAR~\cite{krizhevsky2009learning} and SVHN~\cite{netzer2011reading}, and 5 for ImageNet~\cite{deng2009imagenet}.

\begin{figure*}[t]
	\centering
	\includegraphics[width=0.85\textwidth]{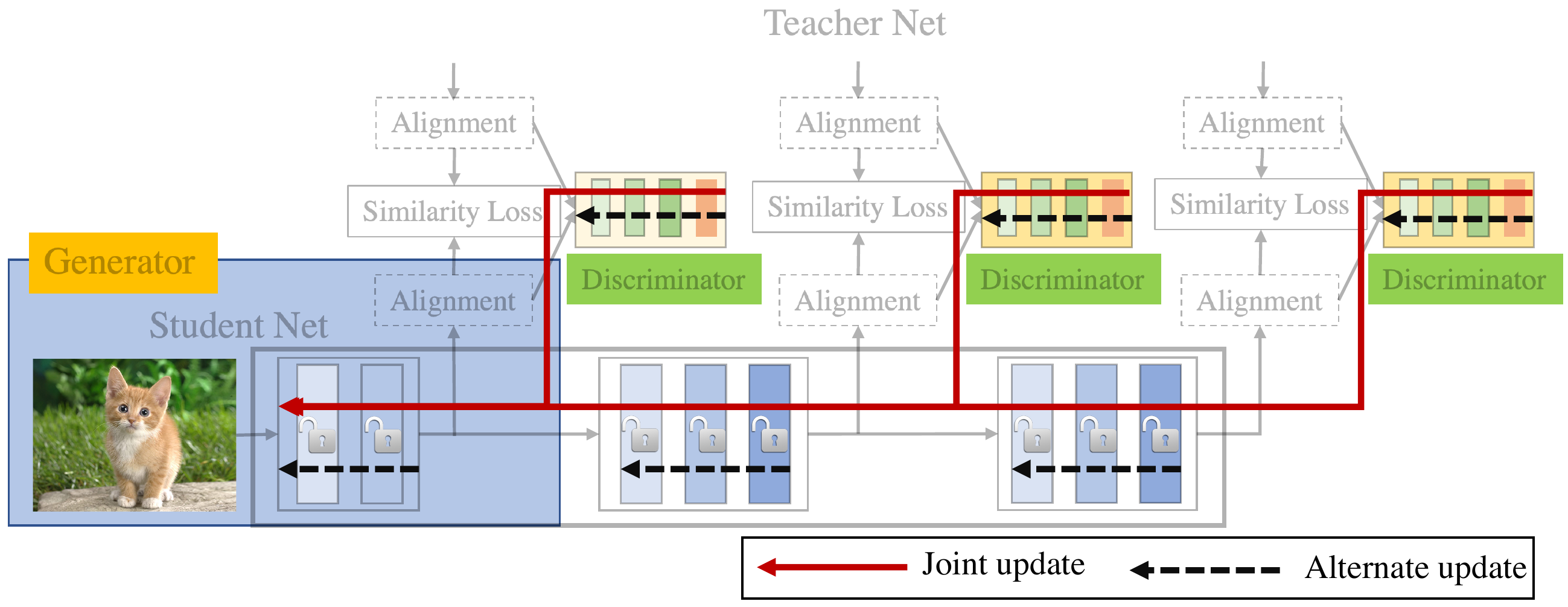}
	\caption{Illustration of our two gradient update strategies. Red line indicates {\em joint training strategy} and black dash line indicates {\em alternate updateing strategy}. More details can be referred to Section~\ref{update_str}.}
	\label{JA}
\end{figure*}

\begin{algorithm}[t]
	\small
	\caption{Learning Strategy of Similarity and Discriminators.}
	\label{strategy}
	
	
	{\bf Strategy 1: Alternately Update Gradients.} 
	
	{\bf Require:} 
	 Following~\cite{goodfellow2014generative}, we also define the number of steps to apply to the discriminator, $k$, as a hyperparameter. $\alpha$ and $\beta$ are the trade-off coefficients.
	 
	\begin{algorithmic}[1]
		\State	for {\em {number of training iterations}} do:
		\State	\ \ \ \ \ for {\em {k steps}} do:
		\State \ \ \ \ \ \ \ \ \ Sample minibatch of {\em m} examples {$X_1$, . . . , $X_m$} from training data through {\em teacher}  as distribution $p_{\mathcal{T}}(\mathcal{T}_\theta(X_i))$ and through {\em student}  as distribution $p_{\mathcal{S}}(\mathcal{S}_\theta(X_i))$.
		\State \ \ \  \ \ Update the $j$-th {\em discriminator} $D$ by ascending its stochastic gradient:
				$$
				\nabla_{\theta_{D}} \frac{1}{m} \sum_{i=1}^{m}\left[\log D\left(\mathcal{T}_\theta(X_i)\right)+\log \left(1-D\left(\mathcal{S}_\theta(X_i)\right)\right)\right]
				$$
		\State \ \ \ \ \  end for
		
	   \State \ \  \ Sample minibatch of {\em m} examples {$X_1$, . . . , $X_m$} from training data through {\em student}  as distribution $p_{\mathcal{S}}(\mathcal{S}_\theta(X_i))$.
	    \State \ \  \  \ \ Update {\em student} by descending its stochastic gradient:
		$$
		\nabla_{{\theta_{\mathcal{S}}}} \frac{1}{m} \sum_{i=1}^{m} [\alpha \LL_{Sim}+\beta \log \left(1-D\left(\mathcal{S}_\theta(X_i)\right)\right)]
		$$
	    
		\State end for
		\end{algorithmic}
	------------------------------------------------------------------------------------
	
		{\bf Strategy 2: Joint Training.} 
		
		
		\begin{algorithmic}[1]
		\State	for {\em {number of training iterations}} do:
		\State \ \ \ \ \ \ Sample minibatch of {\em m} examples {$X_1$, . . . , $X_m$} from training data distribution $p_{data}(X)$.
		\State \ \  \ \  Update the $j$-th {\em discriminator} $D$ and the {\em student} by descending its stochastic gradient:
		$$
			\nabla_{\theta_{D},_{\theta_{\mathcal{S}}}} \frac{1}{m} \sum_{i=1}^{m}\left[\alpha \LL_{Sim}+\beta \LL_{GAN}\right]
		$$
		
%
		
		\State end for
		
		\noindent{The gradient-based updates can use any standard gradient-based learning rule.}
	\end{algorithmic}
\end{algorithm}

\section{Learning Strategy of Similarity and Discriminators} \label{update_str}
\subsection{Joint Training}
For the strategy of joint training, we incorporate the similarity loss in Eq.~\ref{sim} and adversarial loss in Eq.~\ref{gan} into our final loss function based on above definition and analysis. Our whole framework is trained end-to-end by the following objective function: 
\begin{equation}\label{final_loss}
\LL = \alpha \LL_{Sim} + \beta \LL_{GAN} 
\end{equation}

where $\alpha$ and $\beta$ are trade-off weights. We set them as 1 in our experiments by cross validation. We also use the weighted coefficients to balance the contributions of different blocks. For 3-block networks, we ues [0.01, 0.05, 1], and [0.001, 0.01, 0.05, 0.1, 1] for 5-block ones.

\subsection{Alternately Update Gradients}

For alternately updating gradients, we follow the training process of standard generative adversarial networks~\cite{goodfellow2014generative} which trains $D$ to maximize the probability of assigning the correct label to both teacher features and features from student. We simultaneously train $G$ ({\em student}) to minimize $\log (1-D(G(\boldsymbol{x})))$. As in~\cite{goodfellow2014generative}, $D$ and $G$ play the following two-player minimax game with value function $V(G, D)$:
\begin{equation}
	\begin{gathered}
\min _{G} \max _{D} V(D, G)=\mathbb{E}_{\mathcal{T}_\theta(X_i)\sim p_{{ \mathcal{T} }}}[\log D(\mathcal{T}_\theta(X_i))] \hfill \\
	\quad \quad \quad	\quad \quad \quad 	\quad \quad \quad
+\mathbb{E}_{\mathcal{S}_\theta(X_i) \sim p_{\mathcal{S}}}[\log (1-D(\mathcal{S}_\theta(X_i)))]
\end{gathered}
\end{equation}

We update the student network after updating the discriminator in $k$ iterations (we choose $k=1$ in all our experiments). When updating the student network $\mathcal{S}_\theta$, we aim to fool the discriminator by fixing discriminator $D$ and minimizing the similarity loss $\LL_{Sim}$ and GAN loss. More details can be referred in Algorithm~\ref{strategy} and Fig.~\ref{JA}.

The main difference between {\em joint training} and {\em alternate updating} is that in the former strategy, the gradients from discriminators will propagate back to the student backbone, while in the latter strategy, the parameters in discriminators and student will
update separately without any interactions.

\begin{figure*}[t]
	\centering
	\includegraphics[width=0.94\textwidth]{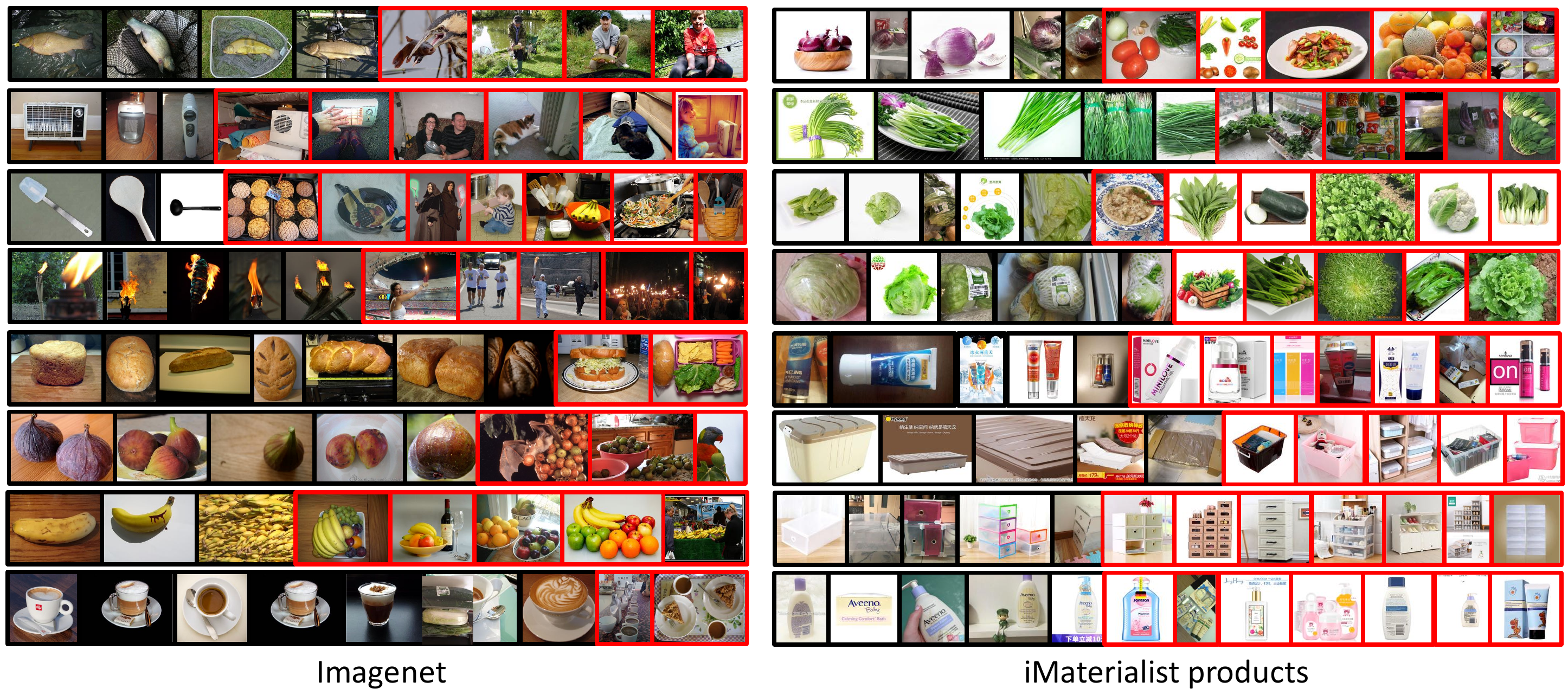}
	\vspace{-0.12in}
	\caption{Image samples from two benchmarks grouped by clean and noisy/multi-label categories. In each group, black box images are
		clean labeled images and red box images are ones with multi-label objects (ImageNet) or noisy label images (iMaterialist products).}
	\label{ap}
\end{figure*}

\section{Multi-Model Ensemble via Adversarial Learning (MEAL)}
We achieve ensemble with a training method that is simple and straight-forward to implement.
As different network structures can obtain different distributions of outputs, which can be viewed as soft labels (knowledge), we adopt these soft labels to train our student, in order to compress knowledge of different architectures into a single network. Thus we can obtain the seemingly contradictory goal of ensembling multiple neural networks at {\em {no additional testing cost}}. 

\begin{algorithm}[h]
	\small
	\caption{Multi-Model Ensemble via Adversarial Learning (MEAL).}
	\label{inf}
	{\bf Stage 1:} 
	
	Building and Pre-training the Teacher Model Zoo $\mathcal{T}=\{\mathcal{T}_\theta^1,\mathcal{T}_\theta^2, \dots \mathcal{T}_\theta^i\}$, including: VGGNet~\cite{simonyan2014very}, ResNet~\cite{he2016deep}, DenseNet~\cite{huang2016densely}, MobileNet~\cite{howard2017mobilenets}, Shake-Shake~\cite{gastaldi2017shake}, etc.
	
	{\bf Stage 2:} 
	\begin{algorithmic}[1]
		\Function{$TSM$}{$\mathcal{T}$}
		\State $\mathcal{T}_\theta \leftarrow RS(\mathcal{T})$ \Comment{Random Selection}
		\State \Return $\mathcal{T}_\theta$
		\EndFunction
		\State	for {\em {each iteration}} do:
		\State \ \ \ \ \ $\mathcal{T}_\theta \leftarrow TSM(\mathcal{T})$
		\Comment{Randomly Select a Teacher Model}
		\State \ \ \ \ \ $\mathcal{S}_\theta = \arg \min _{\mathcal{S}_\theta}\ {\LL}(\mathcal{T}_\theta,\mathcal{S}_\theta)$
		\Comment{Adversarial Learning for a Student}
		\State end for
	\end{algorithmic}
\end{algorithm}

\subsection{Learning Procedure}
To clearly understand what the student learned in our work, we define two conditions. First, the student has the same structure as the teacher network. Second, we choose one structure for student and randomly select a structure for teacher in each iteration as our ensemble learning procedure.

The learning procedure contains two stages. First, we pre-train the teachers to produce a model zoo. Because we use the classification task to train these models, we can use the softmax cross entropy loss as the main training loss in this stage. Second, we minimize the loss function $\LL$ in Eq.~\ref{final_loss} to make the student output similar to that of the teacher output. The learning procedure is explained below in Algorithm~\ref{inf}.

\section{Learning on Noisy Data}

Deep neural networks have achieved notable success in image classification recently due to the collection of massive large-scale labeled datasets such as ImageNet~\cite{deng2009imagenet}, OpenImage~\cite{openimages}, etc. However,  collecting such datasets is time-consuming and expensive, further requires double check from multiple annotators to reduce label error. So a better solution is to automaticly build and learn from an Internet-scale dataset with noisy labels.
Our method is fairly easy to extend to handle noisy data with an automatic way, since the ``soft labels'' predicted from teacher models usually are more accurate than the noisy labels provided by the noisy dataset. We further propose an iterative refinement strategy to boost the performance of our method on noisy labeled dataset.

\subsection{Iterative Refinement}
We propose to use an iterative training strategy to refine the noisy labels, which mainly exhibits three advantages: (1) Rectify sample supervisions with potentially wrong class labels through the teacher model predictions; (2) Improve the quality of predictions from the teacher model $\mathcal{T}$ through the {\em Iterative Refinement}, so that the similarity loss will be more effective; And (3) Mitigate the overfit to noisy smaples when training networks on noisy labeled data, which leads to more robustness of student models against label noise. 

Firstly, we perform an initial training iteration following the method described in Algorithm~\ref{inf}, and obtain a model with the best validation accuracy. This model will be the teacher in the next training iteration. In the second training iteration, we repeat the steps in Algorithm~\ref{inf} with only one change described as follows. We replace the teacher model zoo with the models we learned from the first step. This operation improve the quality of teacher models and promote the teacher model to produce more reliable predictions for student model training, which can improve the quality of student models.

\begin{figure*}[t]
	\centering
	\includegraphics[width=0.243\textwidth]{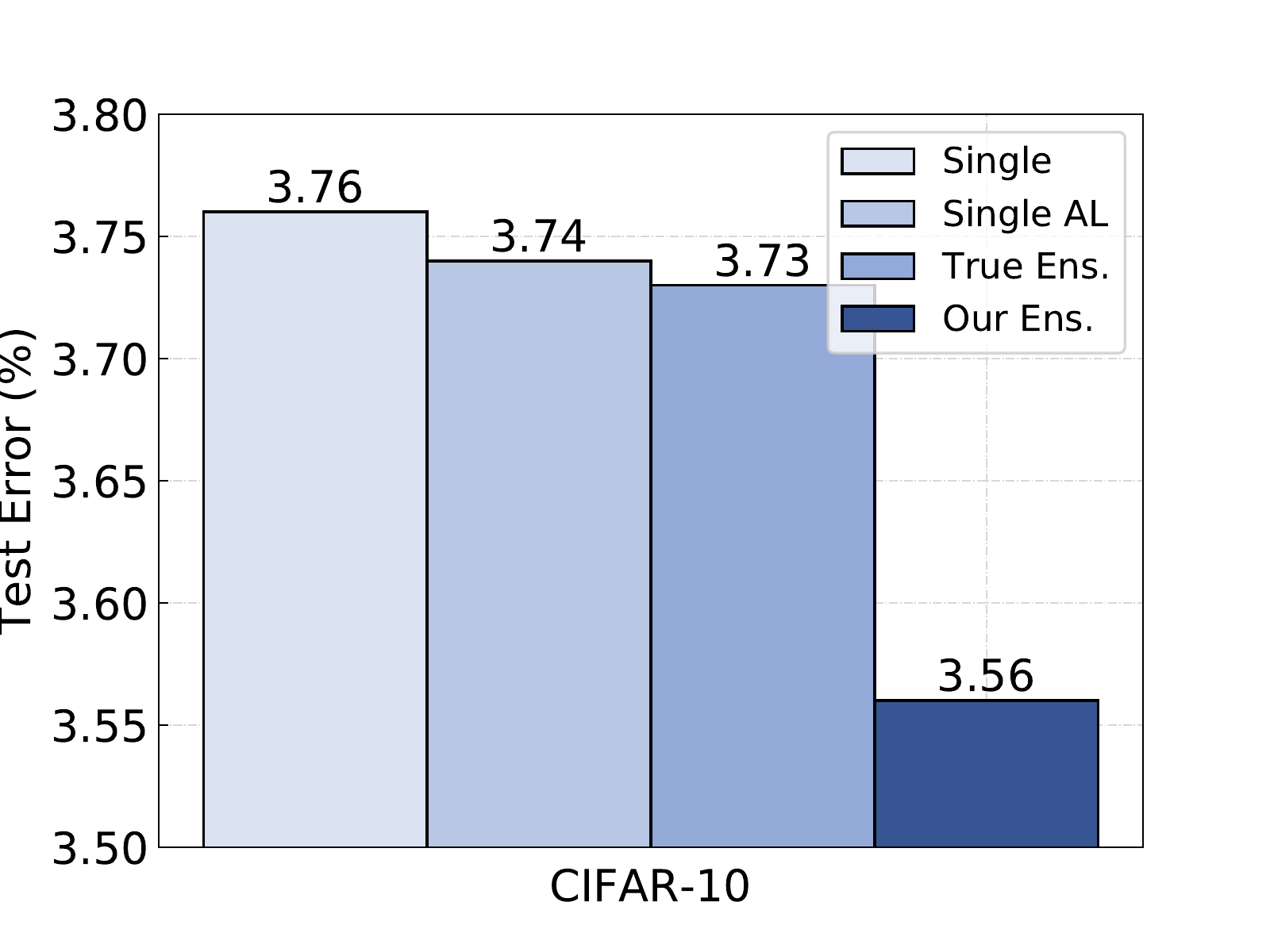}
	\includegraphics[width=0.243\textwidth]{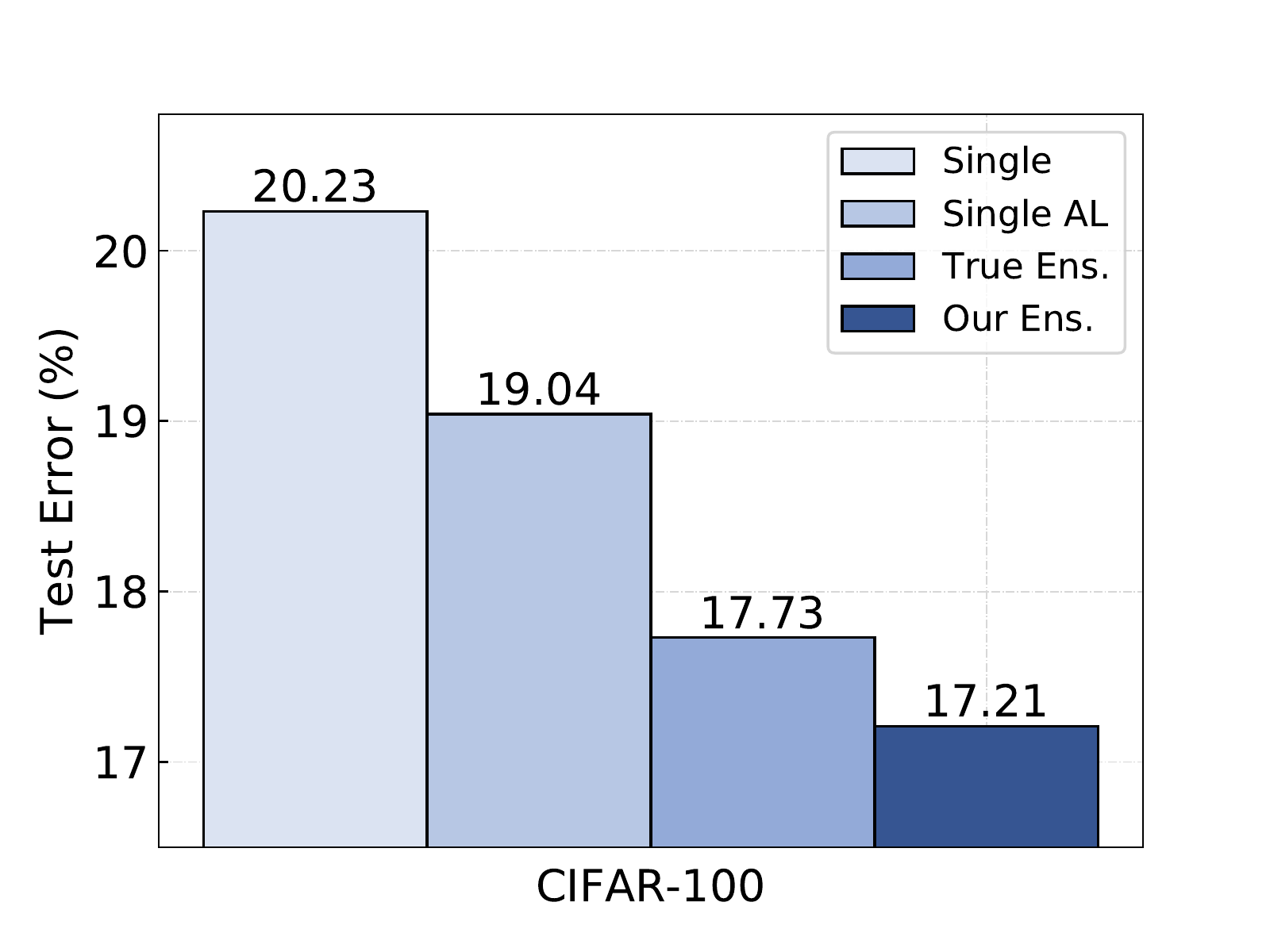}
	\includegraphics[width=0.243\textwidth]{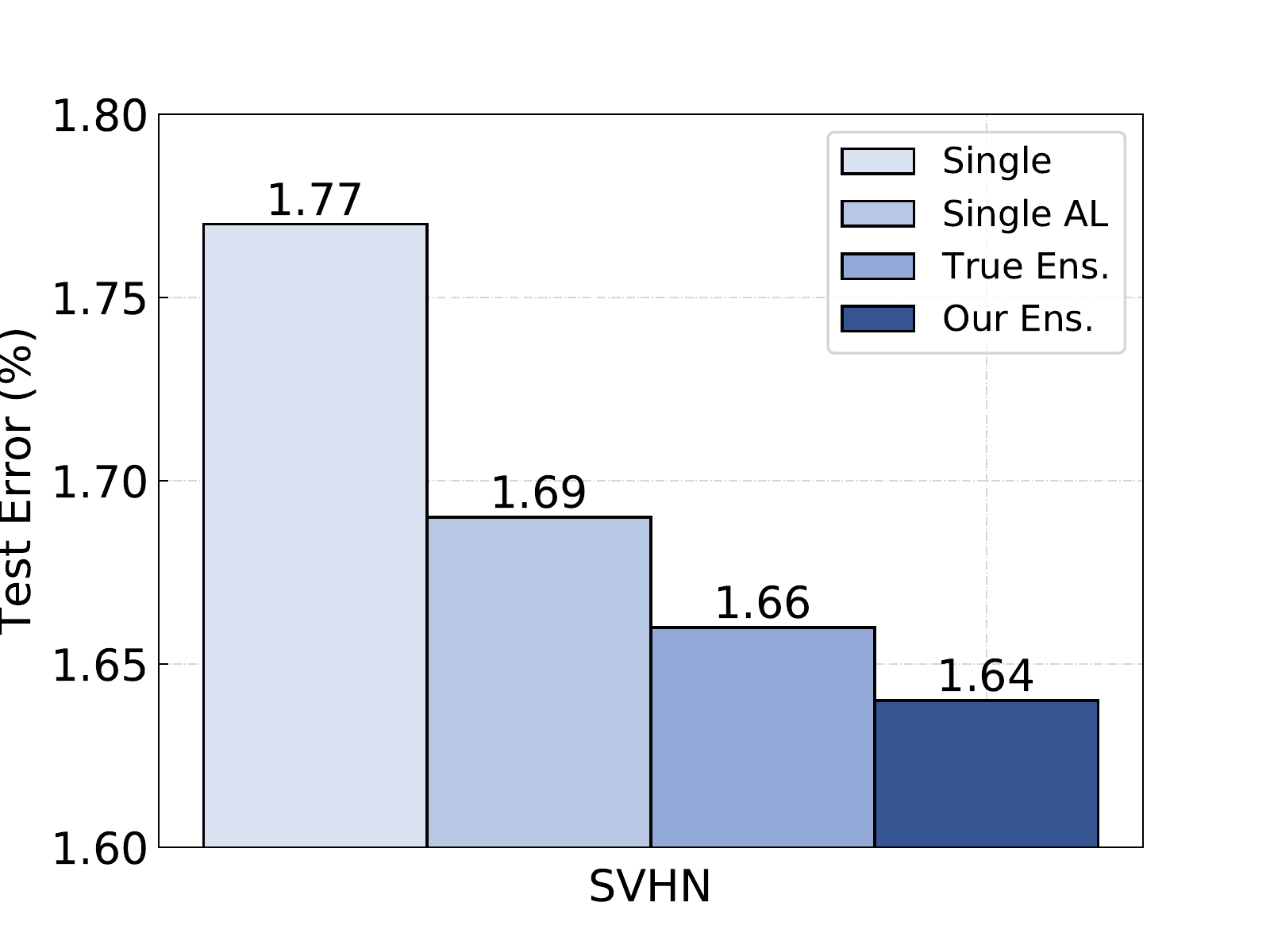}
	\includegraphics[width=0.243\textwidth]{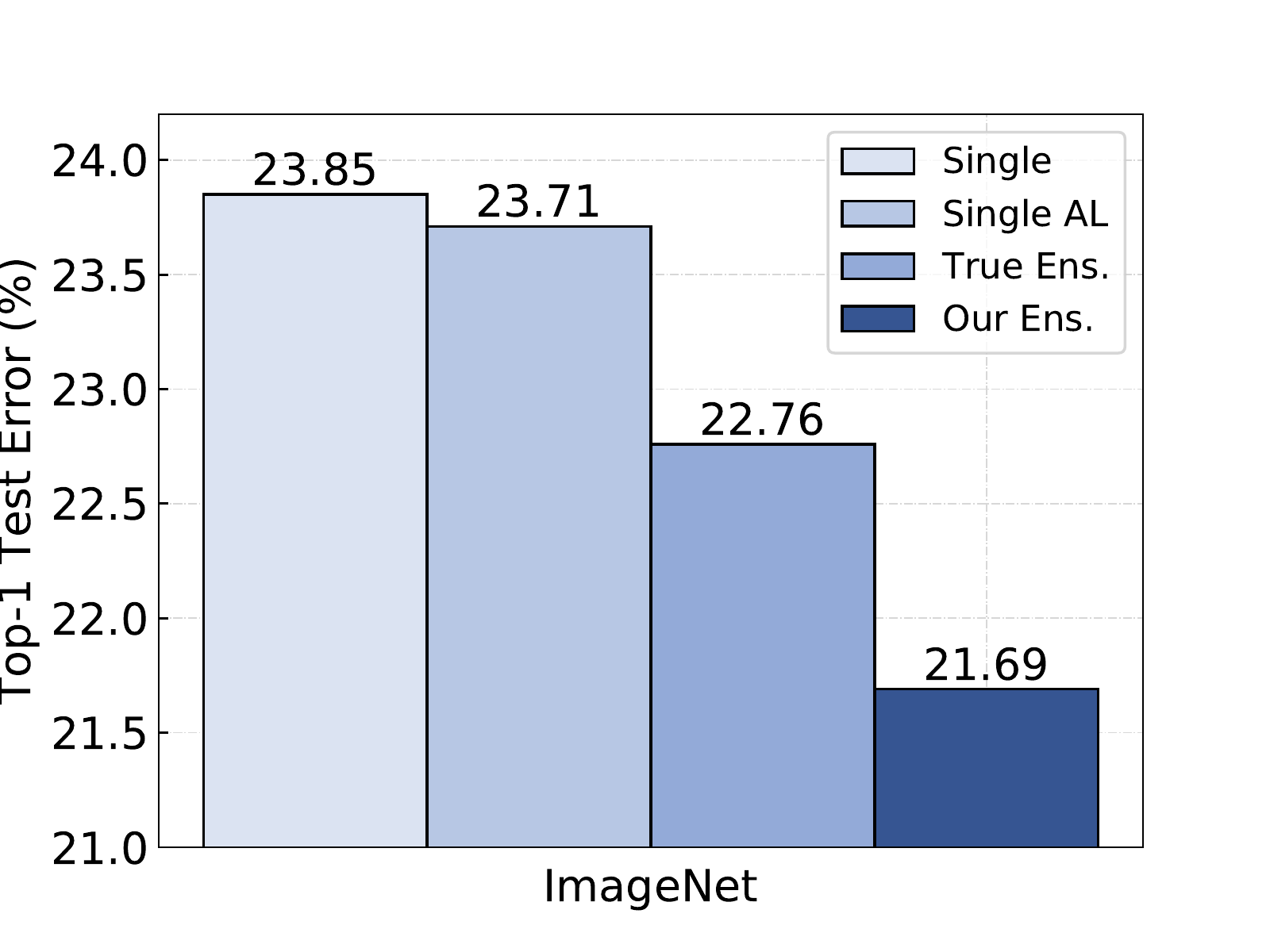}
	\caption{Error rates (\%) on CIFAR-10 and CIFAR-100, SVHN and ImageNet datasets. In each figure, the results from left to right are 1) base model; 2) base model with adversarial learning; 3) true ensemble/traditional ensemble; and 4) our ensemble results. For the first three datasets, we employ DenseNet as student, and ResNet for the last one (ImageNet).}
	\label{com_all}
\end{figure*}

\section{Experiments and Analysis} \label{exp}
We empirically demonstrate the effectiveness of MEAL on several benchmark datasets. We implement our method on the PyTorch~\cite{paszke2017automatic} platform.
\subsection{Datasets}

{\bf CIFAR.} The two CIFAR datasets~\cite{krizhevsky2009learning} consist of colored natural images with a size of 32$\times$32. CIFAR-10 is drawn from 10 and CIFAR-100 is drawn from 100 classes. In each dataset, the train and test sets contain 50,000 and 10,000 images, respectively. A standard data augmentation scheme\footnote{zero-padded with 4 pixels on both sides, randomly cropped to produce 32x32 images, and horizontally mirror with probability 0.5.}~\cite{lee2015deeply,romero2015fitnets,larsson2016fractalnet,huang2017snapshot,liu2017learning} is used.  We report the test errors in this section with training on the whole training set.

\noindent{\textbf{SVHN.}} The Street View House Number (SVHN) dataset~\cite{netzer2011reading} consists of 32$\times$32 colored digit images, with one class for each digit. The train and test sets contain 604,388 and 26,032 images, respectively. Following previous works~\cite{goodfellow2013maxout,huang2016deep,huang2017snapshot,liu2017learning}, we split a subset of 6,000 images for validation, and train on the remaining images without data augmentation.

\noindent{\textbf{ImageNet.}} The ILSVRC 2012 classification dataset~\cite{deng2009imagenet} consists of 1000 classes, with a number of 1.2 million training images and 50,000 validation images. We adopt the the data augmentation scheme following~\cite{krizhevsky2012imagenet} and apply the same operation as ~\cite{huang2017snapshot} at test time.

\noindent{\textbf{iMaterialist Challenge Dataset\footnote{A large-scale, noisy, fine-grained, product classification dataset at FGVC6, CVPR 2019. Website: \url{https://github.com/malongtech/imaterialist-product-2019}.}.}} The iMaterialist Dataset contains about one million product images with 2019 classes for training, about 10K images for validation and 90K images for testing. This dataset is fairly challenging since about 30\% training images are with incorrect labels. We follow the evaluation in the competition which uses top-3 classification error as metric. We also provide top-1 results in our experiments.

\subsection{Networks}
We adopt several popular network architectures as our teacher model zoo, including VGGNet~\cite{simonyan2014very}, ResNet~\cite{he2016deep}, DenseNet~\cite{huang2016densely}, MobileNet~\cite{howard2017mobilenets}, shake-shake~\cite{gastaldi2017shake}, etc. For VGGNet, we use 19-layer with Batch Normalization~\cite{ioffe2015batch}. For ResNet, we use 18-layer network for CIFAR and SVHN and 50-layer for ImagNet. For DenseNet, we use the $BC$ structure with depth L=100, and growth rate k=24. For shake-shake, we use 26-layer 2$\times$96d version. Note that due to the high computing costs, we use shake-shake as a teacher only when the student is shake-shake network.

\subsection{Ablation Studies} \label{ablation_s}
We first investigate each design principle of our MEAL framework with joint training strategy. We design several controlled experiments on CIFAR-10 with VGGNet-19 w/BN (both to teacher and student) for this ablation study. A consistent setting is imposed on all the experiments, unless when some components or structures are examined.

\begin{table}[h]
	\centering
	\caption{{Ablation study on CIFAR-10 using VGGNet-19 w/BN}. Please
		refer to Section~\ref{ablation_s} for more details.}
	\vspace{1.2ex}
	\resizebox{0.48\textwidth}{!}{%
		\label{ablation}
		\begin{tabular}{c|c|c|c|c|c}
			\hline
			$\ell_1$  \bf dis.    &  $\ell_2$ \bf dis. & \bf {Cross-Entropy}    &   \bf  Intermediate   &   \bf Adversarial  & \bf Test Errors (\%) \\ \hline \hline
			\multicolumn{5}{c|}{\textbf{Base Model (VGG-19 w/ BN)~\cite{simonyan2014very}}} &   {\bf 6.34}    \\ \hline\hline
			\Checkmark	&                                    &                        &                                      &                                     &   6.97  \\ \hline
			&         \Checkmark         &                        &                                      &                                     &    6.22  \\ \hline
			&                                    &    \Checkmark  &                                      &                                     &   6.18   \\ \hline
			&                                    &    \Checkmark  &           \Checkmark         &                                     &   6.10   \\ \hline
			&          \Checkmark        &                        &          \Checkmark          &                                     &   6.17    \\ \hline
			&                                    &    \Checkmark  &          \Checkmark          &        \Checkmark           &  \bf 5.83    \\ \hline 
			\Checkmark  &                                    &                        &          \Checkmark          &        \Checkmark           &    7.57 \\ \hline
		\end{tabular}
	}
\end{table}

The results are mainly summarized in Table~\ref{ablation}. The first three rows indicate that we only use $\ell_1$, $\ell_2$ or cross-entropy loss from the last layer of a network. It's similar to the {\em {Knowledge Distillation}} method. We can observe that use cross-entropy achieve the best accuracy. Then we employ more intermediate outputs to calculate the loss, as shown in rows 4 and 5. It's obvious that including more layers improves the performance. Finally, we involve the discriminators to exam the effectiveness of adversarial learning. Using cross-entropy, intermediate layers and adversarial learning achieve the best result. Additionally, we use average based adaptive pooling for alignment. We also tried max operation, the accuracy is much worse (6.32\%).

\subsection{Results of Multi-Model Ensemble}
\noindent{\textbf{Comparison with Different Learning Strategy.}} We compare MEAL with {\em joint update} and {\em alternate update} strategies. The results are shown in Table~\ref{alternate}, we employ several network architectures in this comparison. All models are trained with the same epochs. It can be observed that in the most cases {\em joint training} obtains better performance than {\em alternate update} on all the networks, so in the following experiments, we use {\em joint update} as our basic learning method unless otherwise noted.

\begin{table}[h]
	\centering
	\caption{ {Comparison of error rate (\%) with different learning strategies on CIFAR-10.}}
	\resizebox{0.48\textwidth}{!}{%
		\label{alternate}
		\begin{tabular}{c|c|c}
			\toprule
			\bf Network &  \bf Alternate Updating (A) (\%)  & \bf Joint Training (J)  (\%) \\ \hline\hline
			VGG-19 w/ BN~\cite{simonyan2014very}  & 5.87 & \bf 5.55  \\ \midrule 
			GoogLeNet~\cite{szegedy2015going} &    \bf 4.39  &   4.83 \\ \midrule
			ResNet-18~\cite{he2016deep} &     4.43  &  \bf 4.35 \\ \midrule
			DenseNet-BC ($k$=24)~\cite{huang2016densely}	&  3.99   & \bf 3.54   \\ 
			\bottomrule
		\end{tabular}
	}
\end{table}

\noindent{\textbf{Comparison with Traditional Ensemble.}} The results are summarized in Fig.~\ref{com_all} and Table~\ref{ens}. In Figure~\ref{com_all}, we compare the error rate using the same architecture on a variety of datasets (except ImageNet). It can be observed that our results consistently outperform the single and traditional methods on these datasets. The traditional ensembles are obtained through averaging the final predictions across all teacher models. In Table~\ref{ens}, we compare error rate using different architectures on the same dataset. In most cases, our ensemble method achieves lower error than any of the baselines, including the single model and traditional ensemble.

\begin{table}[h]
	\centering
	\caption{ {Error rate (\%) using different network architectures on CIFAR-10 dataset.}}
	\resizebox{0.48\textwidth}{!}{%
		\label{ens}
		\begin{tabular}{c|c|c|c}
			\toprule
			\bf Network &  \bf Single (\%) & \bf Traditional Ens. (\%) & \bf Our Ens. (\%) \\ \hline\hline
			MobileNet~\cite{howard2017mobilenets}  & 10.70  & -- & \bf 8.09  \\ \midrule 
			VGG-19 w/ BN~\cite{simonyan2014very}  & 6.34  & -- & \bf 5.55  \\ \midrule 
			DenseNet-BC ($k$=24)~\cite{huang2016densely}	&   3.76 &  3.73 & \bf 3.54   \\ \midrule
			Shake-Shake-26 2x96d~\cite{gastaldi2017shake} &  2.86    & 2.79 &  \bf 2.54 \\ 
			\bottomrule
		\end{tabular}
	}
\end{table}

\noindent{\textbf{Comparison with Dropout.}} We compare MEAL with the ``Implicit'' method Dropout~\cite{srivastava2014dropout}. The results are shown in Table~\ref{drop}, we employ several network architectures in this comparison. All models are trained with the same epochs. We use a probability of 0.2 for drop nodes during training. It can be observed that our method achieves better performance than Dropout on all these networks.

\begin{table}[h]
	\centering
	\caption{ {Comparison of error rate (\%) with Dropout~\cite{srivastava2014dropout} baseline on CIFAR-10.}}
	\resizebox{0.43\textwidth}{!}{%
		\label{drop}
		\begin{tabular}{c|c|c}
			\toprule
			\bf Network &  \bf Dropout (\%)  & \bf Our Ens. (\%) \\ \hline\hline
			VGG-19 w/ BN~\cite{simonyan2014very}  &  6.89 & \bf 5.55  \\ \midrule 
			GoogLeNet~\cite{szegedy2015going} &     5.37  &  \bf 4.83 \\ \midrule
			ResNet-18~\cite{he2016deep} &     4.69  &  \bf 4.35 \\ \midrule
			DenseNet-BC ($k$=24)~\cite{huang2016densely}	&  3.75   & \bf 3.54   \\ 
			\bottomrule
		\end{tabular}
	}
\end{table}	

\begin{table}[h]
	\centering
	\caption{\bf {Val. error (\%) on ImageNet dataset.}}
	\resizebox{0.48\textwidth}{!}{%
		\label{imagenet}
		\begin{tabular}{c|c|c|c|c}
			\toprule
			Method & Top-1 (\%) & Top-5 (\%) & \#FLOPs & Inference Time (per/image)\\ \hline\hline
			\multicolumn{5}{c}{\textbf{Teacher Networks:}}     \\ \midrule
			VGG-19 w/BN	&   25.76   & 8.15   &  19.52B & $5.70\times10^{-3}$s \\ \midrule
			ResNet-50	&  23.85    & 7.13   &  4.09B & $1.10\times10^{-2}$s\\ \midrule 
			Ours  (ResNet-50)	&    23.58  &  6.86  &  4.09B  &  $ 1.10\times10^{-2}$s \\ \midrule \midrule
			Traditional Ens.  &  22.76    &  6.49  & 23.61B & $1.67\times10^{-2}$s \\ \midrule
			Ours Plus {\bf J} (ResNet-50)	&   \bf 21.79   & 5.99   &  4.09B  &  $ 1.10\times10^{-2}$s \\ \midrule
			Ours Plus {\bf A} (ResNet-50)	&    22.08   &\bf 5.93   & \bf 4.09B  &  $\bf 1.10\times10^{-2}$s \\ \bottomrule
		\end{tabular}
	}
		\vspace{5ex}
\end{table}	

\begin{figure}[t]
	\centering
	\includegraphics[width=0.48\textwidth]{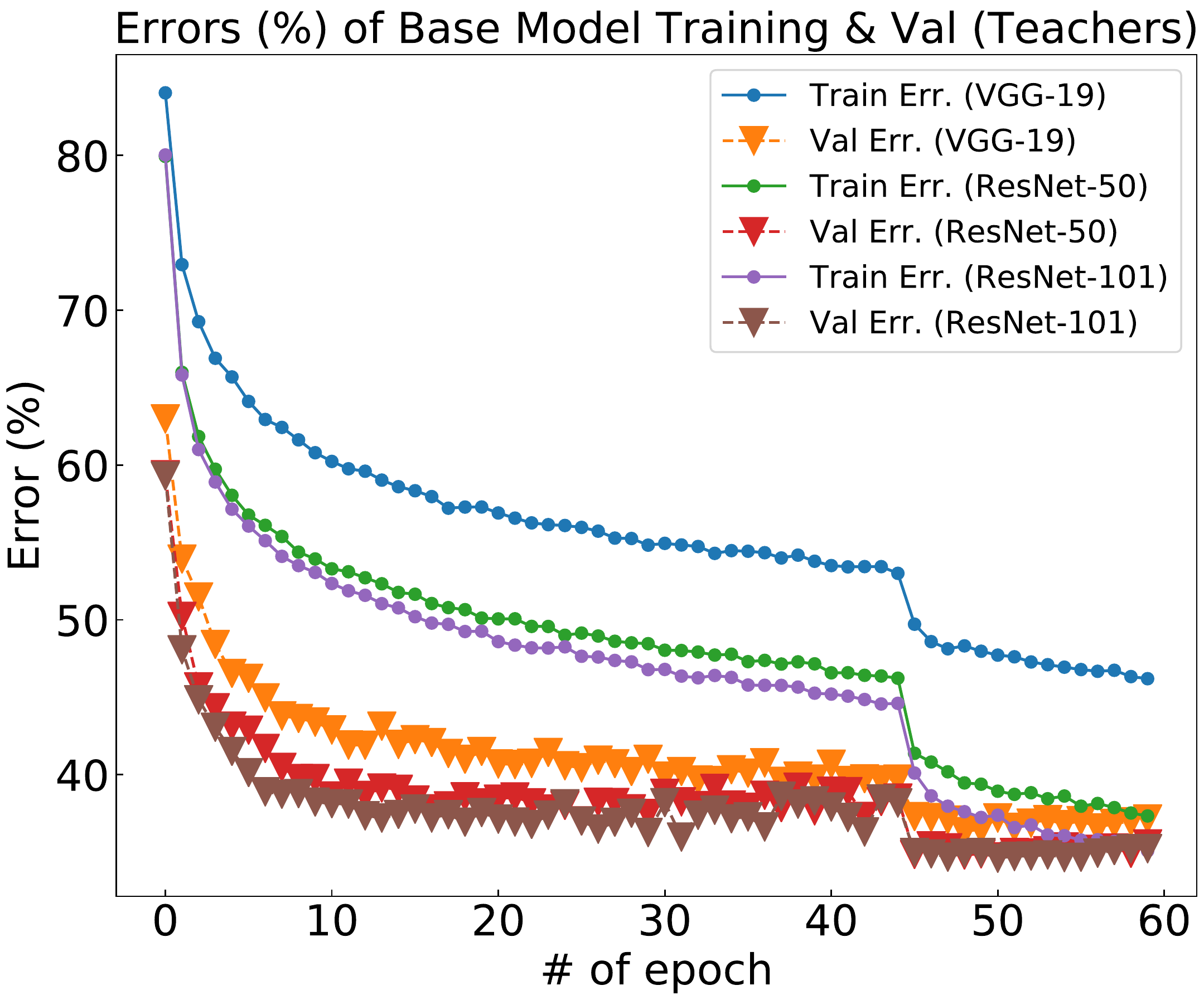}
	\caption{Top-1 error rates (\%) of training and validation with our three base models (VGG-19, ResNet-50 and ResNet-101) on iMaterialist products dataset.}
	\label{base_training}
\end{figure}

\noindent{\textbf{Our Learning-Based Ensemble Results on ImageNet.}} As shown in Table~\ref{imagenet}, we compare our ensemble method with the original model and the traditional ensemble. We use VGG-19 w/BN and ResNet-50 as our teachers, and use ResNet-50 as the student. The \#FLOPs and inference time for traditional ensemble are the sum of individual ones. Therefore, our method has both better performance and higher efficiency. Most notably, our MEAL Plus\footnote{denotes using more powerful teachers like ResNet-101/152.} yields an error rate of Top-1 21.79\%, Top-5 5.99\% on ImageNet, far outperforming the original ResNet-50 23.85\%/7.13\% and the traditional ensemble 22.76\%/6.49\%. This shows great potential on large-scale real-size datasets.

\begin{figure}[t]
	\centering
	\includegraphics[width=0.40\textwidth]{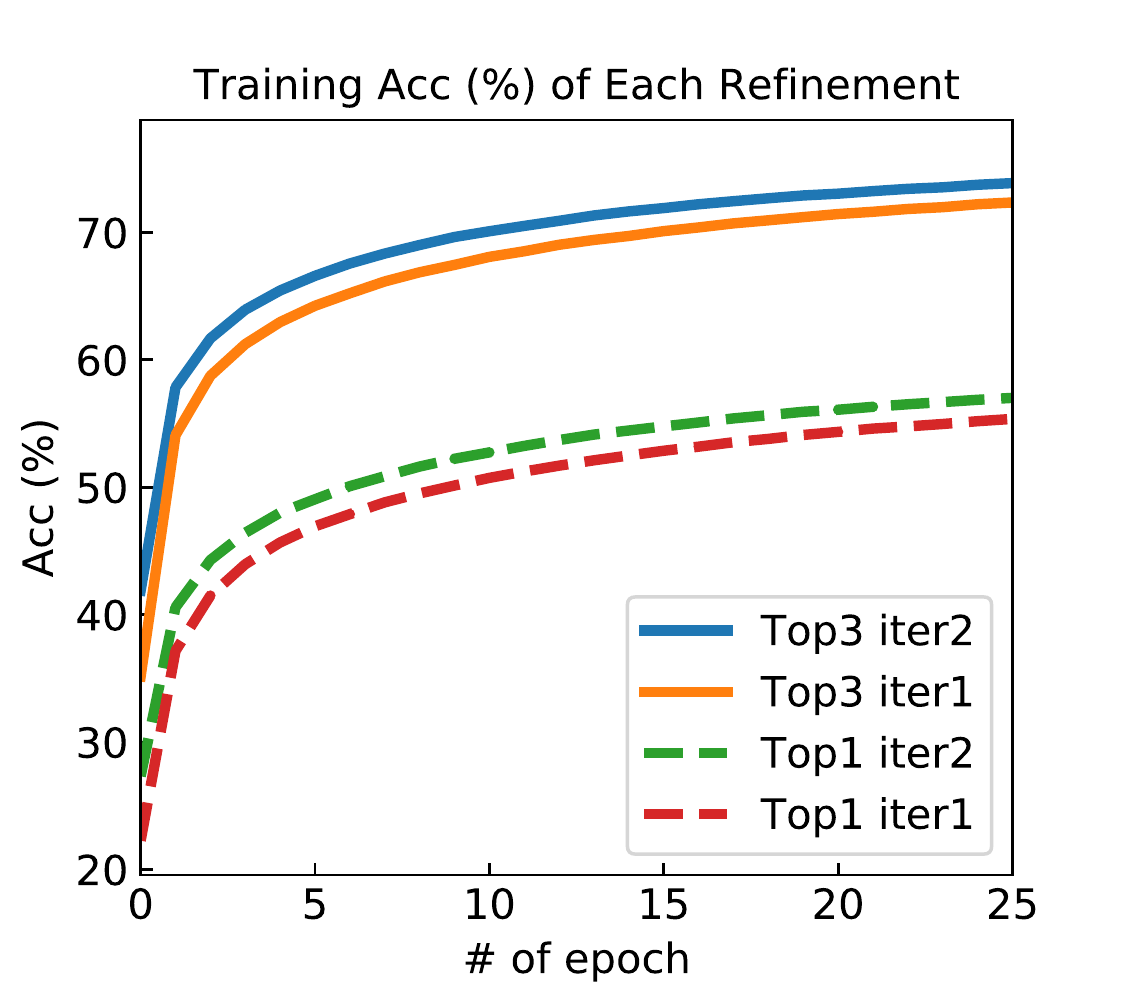}
	\caption{Accuracy curves of our {\em Iterative Refinement} method during training under different re-training budgets on iMaterialist products dataset.}
	\label{re-training}
\end{figure}

\begin{table}[t]
	\centering
	\caption{ {Top-3 error rate (\%)  on iMaterialist products dataset.}}
	\resizebox{0.46\textwidth}{!}{%
		\label{imat}
		\begin{tabular}{c|c|c}
			\toprule
			\bf Network &  \bf Val Set(\%)  & \bf Test Set (\%) \\ \midrule
			VGG-19~\cite{simonyan2014very}  &  11.48 &  10.76 \\ \midrule 
			ResNet-50~\cite{he2016deep}  &  10.03&  9.24  \\ \midrule 
			ResNet-101 (baseline)~\cite{he2016deep}  &  9.19 &  8.96  \\ \midrule 
			\bf MEAL (ResNet-101) &   \bf 8.16 &  \bf 7.81 \\ \midrule \midrule
			MEAL w/ MixUp~\cite{zhang2018mixup} &  --  &  7.57   \\ \midrule
			MEAL w/ CutMix~\cite{yun2019cutmix}	& --  &   7.06  \\ \midrule
			MEAL w/ (CutMix~\cite{yun2019cutmix} + Cosine LR~\cite{loshchilov2016sgdr})& --  & \bf  6.89  \\ 
			\bottomrule
		\end{tabular}
	}
\end{table}	

\begin{figure*}[t]
	\centering
	\includegraphics[width=0.33\textwidth]{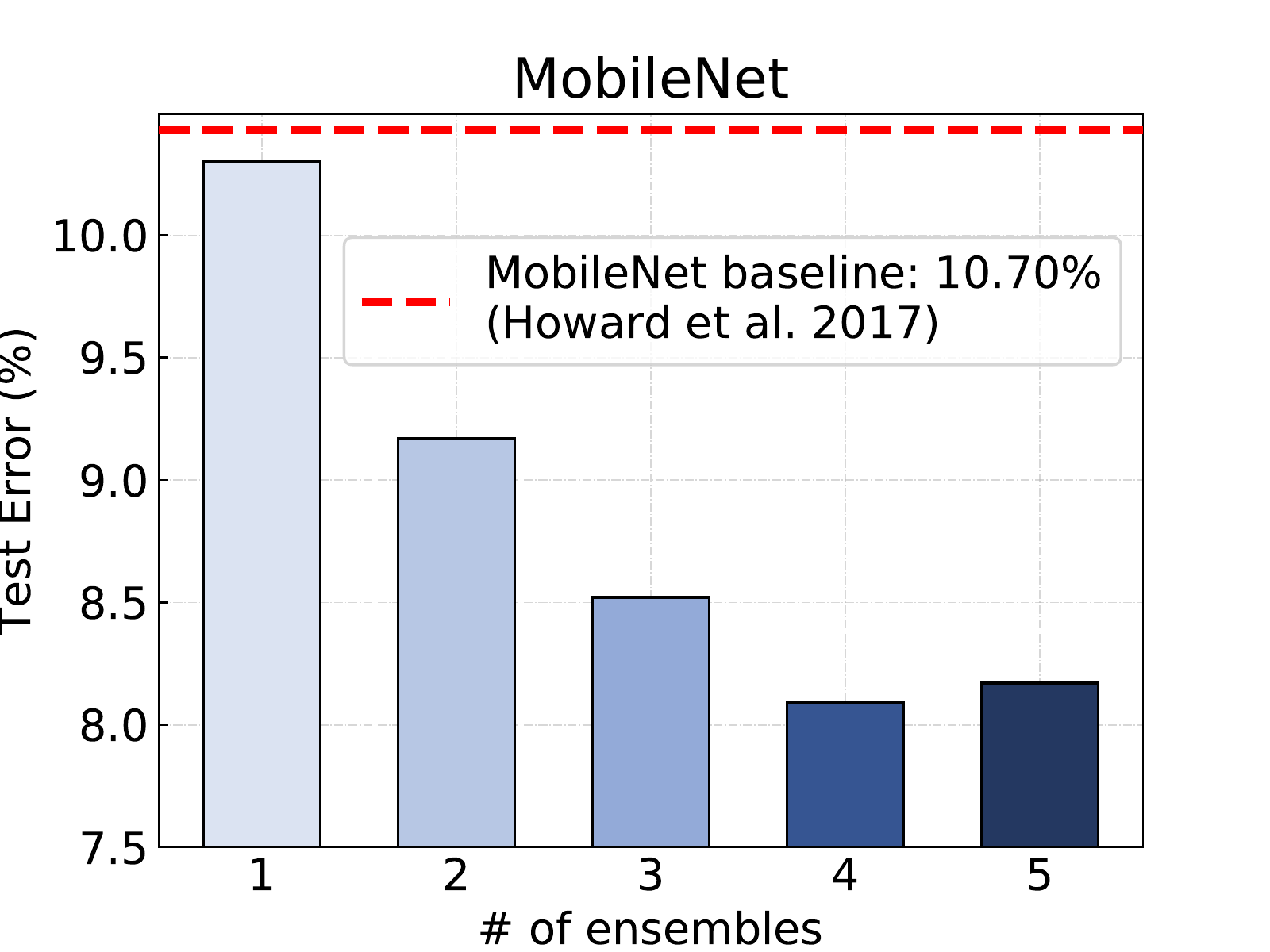}
	\includegraphics[width=0.33\textwidth]{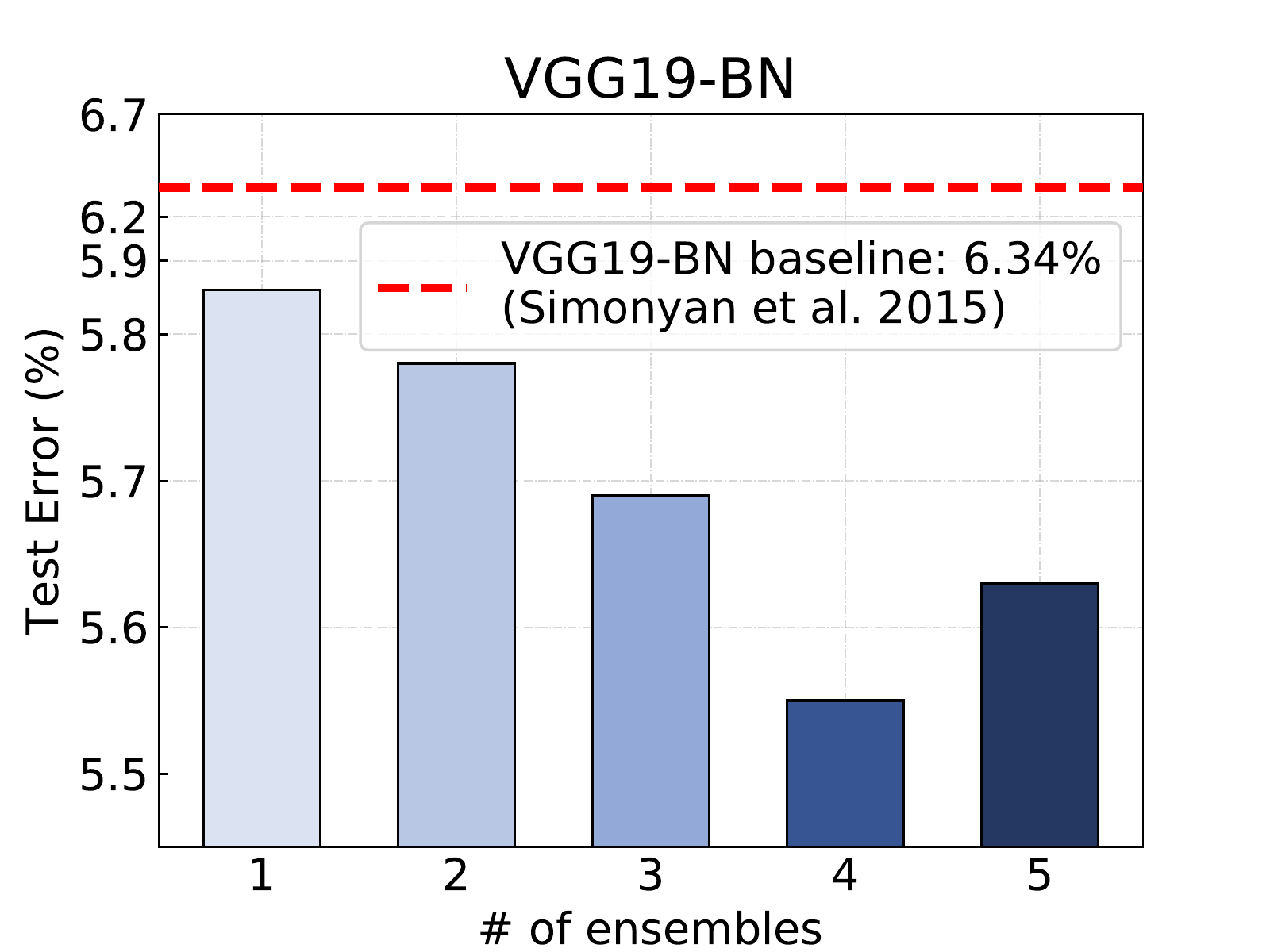}
	\includegraphics[width=0.33\textwidth]{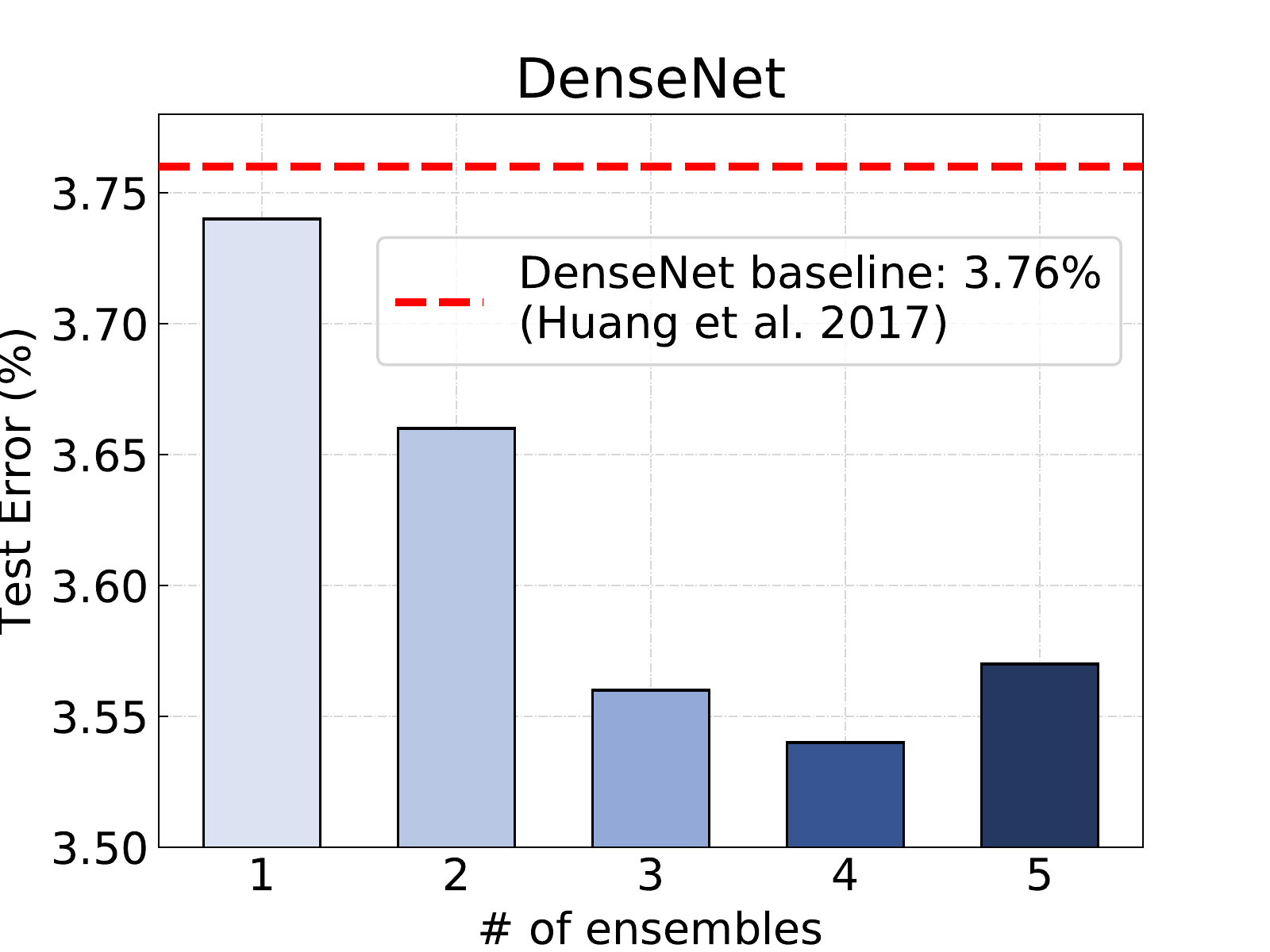}
	\vspace{-0.10in}
	\caption{Error rate (\%) on CIFAR-10 with MobileNet, VGG-19 w/BN and DenseNet.}
	\label{num}
\end{figure*}

\subsection{Results of Noisy Data Refinement}
\noindent{\textbf{Base Model Training (Teachers).}} We first train our base models following the parameter-setting from ImageNet with three network structures: VGG-19, ResNet-50 and ResNet-101. In particular, we use the ImageNet pre-trained networks as initial parameters and the initial learnnig rate is set to 0.01, and then divided by 10 after 45 epochs. The total training budget is 60 epochs. The whole training error curves are illustrated in Fig.~\ref{base_training}. We can observe that because of the large percentage of noisy labels in the training set, the training errors are higher than that on validation set (validation and testing sets are cleaned manually by the organizers). The results on testing set are show in Table~\ref{imat}. The baseline result is 8.96\%, our MEAL outperforms the baseline by 1.15\% (8.96\% {\em vs.} 7.81\%). We further adopt recently proposed data augmentaton method~\cite{zhang2018mixup,yun2019cutmix} to verify whether our model overfits to the training data. From Table~\ref{imat} we can see that after using CutMix~\cite{yun2019cutmix} and Cosine Learning Rate schedule~\cite{loshchilov2016sgdr}, our result further improves to 6.89\%, which demonstrates that our model doesn't overfit to the noisy training data and still has space to improve. 

\noindent{\textbf{{Iterative Refinement.}} Then we iteratively refine our MEAL model with the strategy we described above. Every time we replace the teacher model from the previous round, which can generate better probabilities as supervision for the student model training. We show the train accuracy curves of first and second re-training rounds in Fig.~\ref{re-training}. It is obvious that second re-training has better performance than the first re-training, which verifies the effectiveness of our iterative refinement strategy.

\subsection{Analysis}
\noindent{\textbf{Effectiveness of Ensemble Size.}}
Fig.~\ref{num} displays the performance of three architectures on CIFAR-10 as the ensemble size is varied. Although ensembling more models generally gives better accuracy, we have two important observations. First, we observe that our single model ``ensemble'' already outputs the baseline model with a remarkable margin, which demonstrates the effectiveness of adversarial learning. Second, we observe some drops in accuracy using the VGGNet and DenseNet networks when including too many ensembles for training. In most case, an ensemble of four models obtains the best performance.

\begin{figure}[t]
	\centering
	\includegraphics[width=0.40\textwidth]{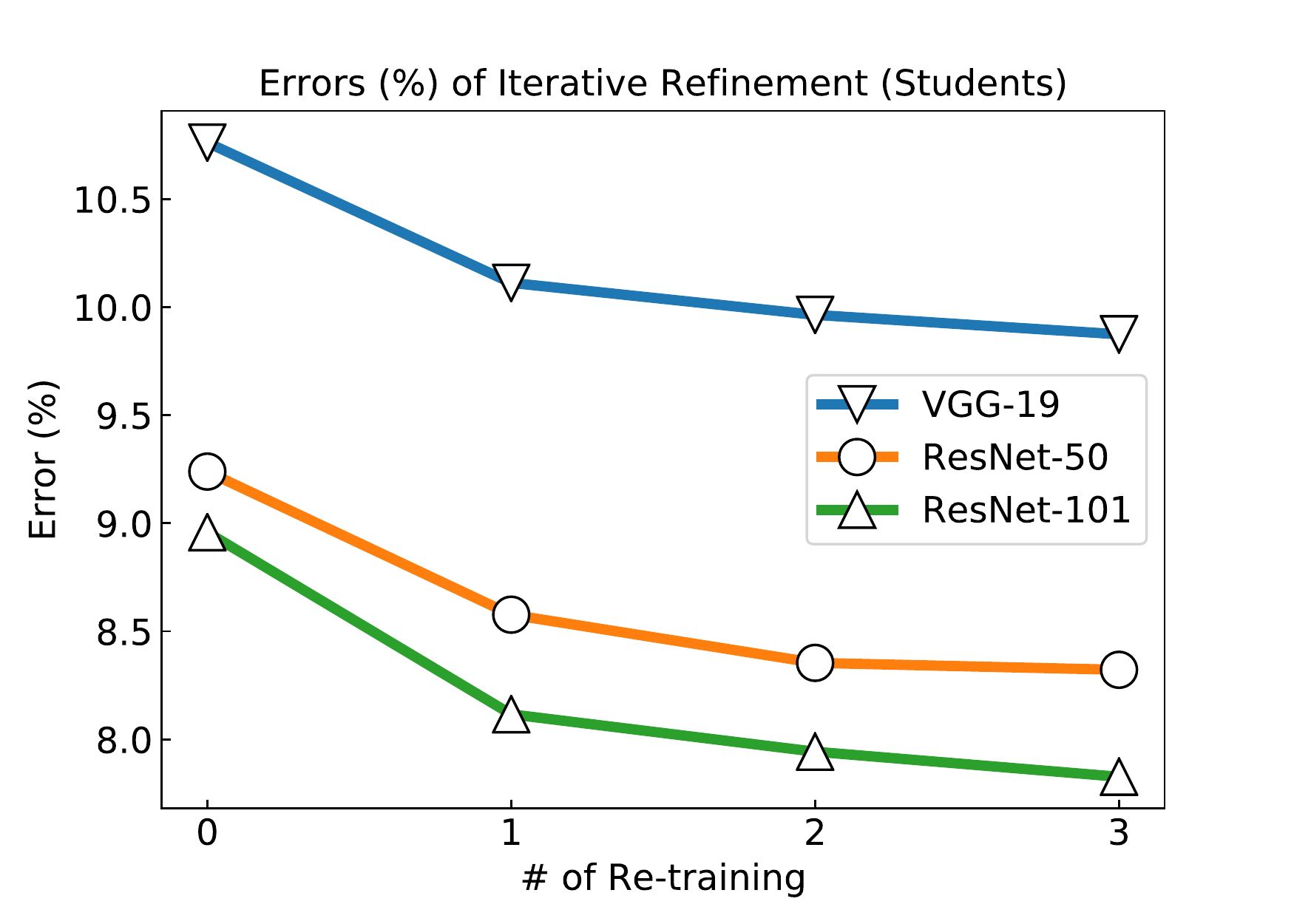}
	\caption{Error rates (\%)  of different re-training time with our three base models (VGG-19, ResNet-50 and ResNet-101) on iMaterialist products testing set. ``0'' indicates the base model performance (teachers).}
	\label{verify}
\end{figure}

\begin{figure}[t]
	\centering
	\includegraphics[width=0.40\textwidth]{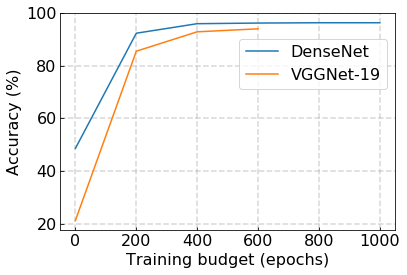}
	\caption{Accuracy of our ensemble method under different training budgets on CIFAR-10.}
	\label{budget}
	\vspace{0.1in}
\end{figure}

\noindent{\textbf{Budget for Training.}}
On CIFAR datasets, the standard training budget is 300 epochs. Intuitively, our ensemble method can benefit from more training budget, since we use the diverse soft distributions as labels. Fig.~\ref{budget} displays the relation between performance and training budget. It appears that more than 400 epochs is the optimal choice and our model will fully converge at about 500 epochs.

\noindent{\textbf{Effectiveness of Re-training Number.}}
Fig.~\ref{verify} displays the performance of three networks on iMaterialist products dataset as the number of re-training is varied. We can observe that the first re-training process generally gives most improvement on all three networks. After that, continuing re-training the models provides very limited boost, but still can increase the accuracy.

\begin{figure}[t]
	\begin{subfigure}[b]{.49\linewidth}
		\centering
		\includegraphics[width=0.98\textwidth]{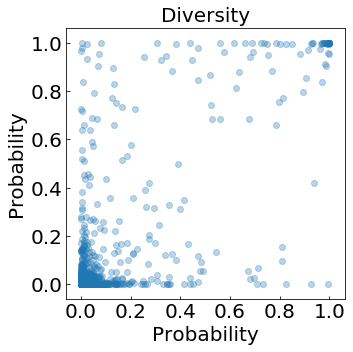}
		\caption{SequeezeNet {\em{vs.}} VGGNet}\label{fig1a}
	\end{subfigure}\hfill
	\begin{subfigure}[b]{.49\linewidth}
		\centering
		\includegraphics[width=0.98\textwidth]{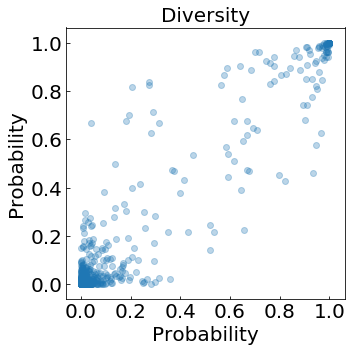}
		\caption{ResNet {\em{vs.}} DenseNet}\label{fig1a}
	\end{subfigure}\hfill
	\vspace{0.1in}
	\begin{subfigure}[b]{.49\linewidth}
		\centering
		\includegraphics[width=0.98\textwidth]{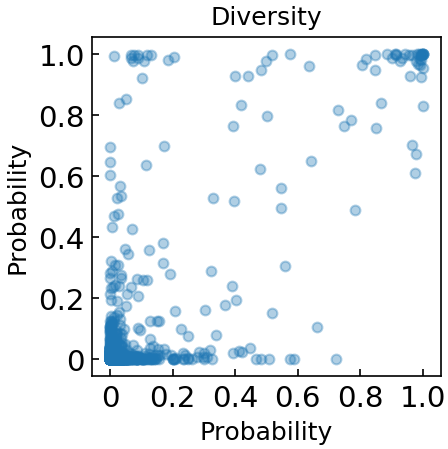}
		\caption{AlexNet {\em vs.} VGGNet}\label{fig1a}
	\end{subfigure}\hfill
	\begin{subfigure}[b]{.49\linewidth}
		\centering
		\includegraphics[width=0.98\textwidth]{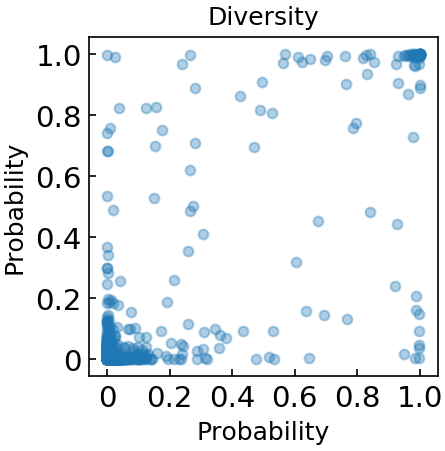}
		\caption{VGGNet {\em vs.} ResNet}\label{fig1a}
	\end{subfigure}\hfill

	\caption{Probability Distributions between five networks.} 
	\label{bubble}
\end{figure}

\noindent{\textbf{Diversity of Supervision.}}
We hypothesize that different architectures create soft labels which are not only informative but also diverse with respect to object categories. We qualitatively measure this diversity by visualizing  the pairwise correlation of softmax outputs from two different networks. To do so, we compute the  softmax predictions for each training image in ImageNet dataset and visualize each pair of the corresponding ones. Fig.~\ref{bubble} displays the bubble maps of four architectures. In the top-left figure, the coordinate of each bubble is a pair of $k$-th predictions ($p^k_{SequeezeNet},p^k_{VGGNet}$), $k=1,2,\dots,1000$, and the top-right figure is ($p^k_{ResNet},p^k_{DenseNet}$). If the label distributions are identical from two networks, the bubbles will be placed on the master diagonal. 
It's very interesting to observe that the top-left (weaker network pairs) has bigger diversity than the top-right (stronger network pairs). It makes sense because the stronger models generally tend to generate predictions close to the ground-truth. In brief, these differences in predictions can be exploited to create effective ensembles and our method is capable of improving the competitive baselines using this kind of diverse supervisions.

\section{Conclusion}
We have presented MEAL, a learning-based ensemble method that can compress multi-model knowledge into a single network with adversarial learning. Our experimental evaluation on three benchmarks CIFAR-10/100, SVHN, ImageNet and iMaterialist Products Dataset verified the effectiveness of our proposed method, which achieved the state-of-the-art accuracy for a variety of network architectures. Our further work will focus on adopting MEAL for cross-domain ensemble and adaption.


%

%
%
%


\ifCLASSOPTIONcaptionsoff
  \newpage
\fi



%

\bibliographystyle{IEEEtran}
\bibliography{mybib}

\begin{thebibliography}{10}
\providecommand{\url}[1]{#1}
\csname url@samestyle\endcsname
\providecommand{\newblock}{\relax}
\providecommand{\bibinfo}[2]{#2}
\providecommand{\BIBentrySTDinterwordspacing}{\spaceskip=0pt\relax}
\providecommand{\BIBentryALTinterwordstretchfactor}{4}
\providecommand{\BIBentryALTinterwordspacing}{\spaceskip=\fontdimen2\font plus
\BIBentryALTinterwordstretchfactor\fontdimen3\font minus
  \fontdimen4\font\relax}
\providecommand{\BIBforeignlanguage}[2]{{%
\expandafter\ifx\csname l@#1\endcsname\relax
\typeout{** WARNING: IEEEtran.bst: No hyphenation pattern has been}%
\typeout{** loaded for the language `#1'. Using the pattern for}%
\typeout{** the default language instead.}%
\else
\language=\csname l@#1\endcsname
\fi
#2}}
\providecommand{\BIBdecl}{\relax}
\BIBdecl

\bibitem{deng2009imagenet}
J.~Deng, W.~Dong, R.~Socher, L.-J. Li, K.~Li, and L.~Fei-Fei, ``Imagenet: A
  large-scale hierarchical image database,'' in \emph{2009 IEEE conference on
  computer vision and pattern recognition}.\hskip 1em plus 0.5em minus
  0.4em\relax Ieee, 2009, pp. 248--255.

\bibitem{openimages}
I.~Krasin, T.~Duerig, N.~Alldrin, A.~Veit \emph{et~al.}, ``Openimages: A public
  dataset for large-scale multi-label and multi-class image classification.''
  \emph{https://github.com/openimages}, 2016.

\bibitem{srivastava2014dropout}
N.~Srivastava, G.~E. Hinton, A.~Krizhevsky \emph{et~al.}, ``Dropout: a simple
  way to prevent neural networks from overfitting.'' \emph{JMLR}, 2014.

\bibitem{wan2013regularization}
L.~Wan, M.~Zeiler, S.~Zhang, Y.~Le~Cun, and R.~Fergus, ``Regularization of
  neural networks using dropconnect,'' in \emph{International conference on
  machine learning}, 2013, pp. 1058--1066.

\bibitem{huang2016deep}
G.~Huang, Y.~Sun, Z.~Liu, D.~Sedra, and K.~Q. Weinberger, ``Deep networks with
  stochastic depth,'' in \emph{European conference on computer vision}.\hskip
  1em plus 0.5em minus 0.4em\relax Springer, 2016, pp. 646--661.

\bibitem{singh2016swapout}
S.~Singh, D.~Hoiem, and D.~Forsyth, ``Swapout: Learning an ensemble of deep
  architectures,'' in \emph{Advances in neural information processing systems},
  2016, pp. 28--36.

\bibitem{maaten2008visualizing}
L.~v.~d. Maaten and G.~Hinton, ``Visualizing data using t-sne,'' \emph{Journal
  of machine learning research}, vol.~9, no. Nov, pp. 2579--2605, 2008.

\bibitem{gastaldi2017shake}
X.~Gastaldi, ``Shake-shake regularization,'' \emph{arXiv preprint
  arXiv:1705.07485}, 2017.

\bibitem{shen2019MEAL}
Z.~Shen, Z.~He, and X.~Xue, ``Meal: Multi-model ensemble via adversarial
  learning,'' in \emph{AAAI}, 2019.

\bibitem{huang2017snapshot}
G.~Huang, Y.~Li, G.~Pleiss, Z.~Liu, J.~E. Hopcroft, and K.~Q. Weinberger,
  ``Snapshot ensembles: Train 1, get m for free,'' in \emph{ICLR}, 2017.

\bibitem{hansen1990neural}
L.~K. Hansen and P.~Salamon, ``Neural network ensembles,'' \emph{IEEE
  transactions on pattern analysis and machine intelligence}, vol.~12, no.~10,
  pp. 993--1001, 1990.

\bibitem{perrone1995networks}
M.~P. Perrone and L.~N. Cooper, ``When networks disagree: Ensemble methods for
  hybrid neural networks,'' in \emph{How We Learn; How We Remember: Toward an
  Understanding of Brain and Neural Systems: Selected Papers of Leon N
  Cooper}.\hskip 1em plus 0.5em minus 0.4em\relax World Scientific, 1995, pp.
  342--358.

\bibitem{krogh1995neural}
A.~Krogh and J.~Vedelsby, ``Neural network ensembles, cross validation, and
  active learning,'' in \emph{Advances in neural information processing
  systems}, 1995, pp. 231--238.

\bibitem{dietterich2000ensemble}
T.~G. Dietterich, ``Ensemble methods in machine learning,'' in
  \emph{International workshop on multiple classifier systems}, 2000, pp.
  1--15.

\bibitem{lakshminarayanan2017simple}
B.~Lakshminarayanan, A.~Pritzel, and C.~Blundell, ``Simple and scalable
  predictive uncertainty estimation using deep ensembles,'' in \emph{Advances
  in Neural Information Processing Systems}, 2017, pp. 6402--6413.

\bibitem{zhu2018knowledge}
X.~Zhu, S.~Gong \emph{et~al.}, ``Knowledge distillation by on-the-fly native
  ensemble,'' in \emph{Advances in Neural Information Processing Systems},
  2018, pp. 7517--7527.

\bibitem{zhou2018diverse}
T.~Zhou, S.~Wang, and J.~A. Bilmes, ``Diverse ensemble evolution: Curriculum
  data-model marriage,'' in \emph{Advances in Neural Information Processing
  Systems}, 2018, pp. 5905--5916.

\bibitem{hinton2015distilling}
G.~Hinton, O.~Vinyals, and J.~Dean, ``Distilling the knowledge in a neural
  network,'' \emph{arXiv preprint arXiv:1503.02531}, 2015.

\bibitem{papernot2016semi}
N.~Papernot, M.~Abadi, U.~Erlingsson, I.~Goodfellow, and K.~Talwar,
  ``Semi-supervised knowledge transfer for deep learning from private training
  data,'' in \emph{ICLR}, 2017.

\bibitem{li2017learning}
Y.~Li, J.~Yang, Y.~Song, L.~Cao, J.~Luo, and L.-J. Li, ``Learning from noisy
  labels with distillation,'' in \emph{Proceedings of the IEEE International
  Conference on Computer Vision}, 2017, pp. 1910--1918.

\bibitem{yim2017gift}
J.~Yim, D.~Joo, J.~Bae, and J.~Kim, ``A gift from knowledge distillation: Fast
  optimization, network minimization and transfer learning,'' in
  \emph{Proceedings of the IEEE Conference on Computer Vision and Pattern
  Recognition}, 2017, pp. 4133--4141.

\bibitem{goodfellow2014generative}
I.~Goodfellow, J.~Pouget-Abadie, M.~Mirza, B.~Xu, D.~Warde-Farley, S.~Ozair,
  A.~Courville, and Y.~Bengio, ``Generative adversarial nets,'' in
  \emph{Advances in neural information processing systems}, 2014, pp.
  2672--2680.

\bibitem{bagherinezhad2018label}
H.~Bagherinezhad, M.~Horton, M.~Rastegari, and A.~Farhadi, ``Label refinery:
  Improving imagenet classification through label progression,'' in
  \emph{ECCV}, 2018.

\bibitem{arjovsky2017wasserstein}
M.~Arjovsky, S.~Chintala, and L.~Bottou, ``Wasserstein generative adversarial
  networks,'' in \emph{International conference on machine learning}, 2017, pp.
  214--223.

\bibitem{gulrajani2017improved}
I.~Gulrajani, F.~Ahmed, M.~Arjovsky, V.~Dumoulin, and A.~C. Courville,
  ``Improved training of wasserstein gans,'' in \emph{Advances in neural
  information processing systems}, 2017, pp. 5767--5777.

\bibitem{kodali2017convergence}
N.~Kodali, J.~Abernethy, J.~Hays, and Z.~Kira, ``On convergence and stability
  of gans,'' \emph{arXiv preprint arXiv:1705.07215}, 2017.

\bibitem{fedus2017many}
W.~Fedus, M.~Rosca, B.~Lakshminarayanan, A.~M. Dai, S.~Mohamed, and
  I.~Goodfellow, ``Many paths to equilibrium: Gans do not need to decrease a
  divergence at every step,'' \emph{arXiv preprint arXiv:1710.08446}, 2017.

\bibitem{mao2017least}
X.~Mao, Q.~Li, H.~Xie, R.~Y. Lau, Z.~Wang, and S.~Paul~Smolley, ``Least squares
  generative adversarial networks,'' in \emph{Proceedings of the IEEE
  International Conference on Computer Vision}, 2017, pp. 2794--2802.

\bibitem{isola2017image}
P.~Isola, J.-Y. Zhu, T.~Zhou, and A.~A. Efros, ``Image-to-image translation
  with conditional adversarial networks,'' in \emph{Proceedings of the IEEE
  conference on computer vision and pattern recognition}, 2017, pp. 1125--1134.

\bibitem{zhu2017unpaired}
J.-Y. Zhu, T.~Park, P.~Isola, and A.~A. Efros, ``Unpaired image-to-image
  translation using cycle-consistent adversarial networks,'' in
  \emph{Proceedings of the IEEE international conference on computer vision},
  2017, pp. 2223--2232.

\bibitem{zhu2017toward}
J.-Y. Zhu, R.~Zhang, D.~Pathak, T.~Darrell, A.~A. Efros, O.~Wang, and
  E.~Shechtman, ``Toward multimodal image-to-image translation,'' in
  \emph{Advances in Neural Information Processing Systems}, 2017, pp. 465--476.

\bibitem{liu2017unsupervised}
M.-Y. Liu, T.~Breuel, and J.~Kautz, ``Unsupervised image-to-image translation
  networks,'' in \emph{Advances in neural information processing systems},
  2017, pp. 700--708.

\bibitem{huang2018multimodal}
X.~Huang, M.-Y. Liu, S.~Belongie, and J.~Kautz, ``Multimodal unsupervised
  image-to-image translation,'' in \emph{Proceedings of the European Conference
  on Computer Vision (ECCV)}, 2018, pp. 172--189.

\bibitem{shen2019towards}
Z.~Shen, M.~Huang, J.~Shi, X.~Xue, and T.~Huang, ``Towards instance-level
  image-to-image translation,'' in \emph{CVPR}, 2019.

\bibitem{johnson2018image}
J.~Johnson, A.~Gupta, and L.~Fei-Fei, ``Image generation from scene graphs,''
  in \emph{Proceedings of the IEEE Conference on Computer Vision and Pattern
  Recognition}, 2018, pp. 1219--1228.

\bibitem{bai2018finding}
Y.~Bai, Y.~Zhang, M.~Ding, and B.~Ghanem, ``Finding tiny faces in the wild with
  generative adversarial network,'' pp. 21--30, 2018.

\bibitem{chen2015net2net}
T.~Chen, I.~Goodfellow, and J.~Shlens, ``Net2net: Accelerating learning via
  knowledge transfer,'' in \emph{ICLR}, 2016.

\bibitem{xu2017training}
Z.~Xu, Y.-C. Hsu, and J.~Huang, ``Training shallow and thin networks for
  acceleration via knowledge distillation with conditional adversarial
  networks,'' \emph{arXiv preprint arXiv:1709.00513}, 2017.

\bibitem{tarvainen2017mean}
A.~Tarvainen and H.~Valpola, ``Mean teachers are better role models:
  Weight-averaged consistency targets improve semi-supervised deep learning
  results,'' in \emph{Advances in neural information processing systems}, 2017,
  pp. 1195--1204.

\bibitem{anil2018large}
R.~Anil, G.~Pereyra, A.~Passos, R.~Ormandi, G.~E. Dahl, and G.~E. Hinton,
  ``Large scale distributed neural network training through online
  distillation,'' in \emph{ICLR}, 2018.

\bibitem{chen2018darkrank}
Y.~Chen, N.~Wang, and Z.~Zhang, ``Darkrank: Accelerating deep metric learning
  via cross sample similarities transfer,'' in \emph{Thirty-Second AAAI
  Conference on Artificial Intelligence}, 2018.

\bibitem{park2019relational}
W.~Park, D.~Kim, Y.~Lu, and M.~Cho, ``Relational knowledge distillation,'' in
  \emph{Proceedings of the IEEE Conference on Computer Vision and Pattern
  Recognition}, 2019, pp. 3967--3976.

\bibitem{heo2018knowledge}
B.~Heo, M.~Lee, S.~Yun, and J.~Y. Choi, ``Knowledge transfer via distillation
  of activation boundaries formed by hidden neurons,'' \emph{arXiv preprint
  arXiv:1811.03233}, 2018.

\bibitem{furlanello2018born}
T.~Furlanello, Z.~C. Lipton, M.~Tschannen, L.~Itti, and A.~Anandkumar, ``Born
  again neural networks,'' \emph{arXiv preprint arXiv:1805.04770}, 2018.

\bibitem{uijlings2018revisiting}
J.~Uijlings, S.~Popov, and V.~Ferrari, ``Revisiting knowledge transfer for
  training object class detectors,'' in \emph{Proceedings of the IEEE
  Conference on Computer Vision and Pattern Recognition}, 2018, pp. 1101--1110.

\bibitem{crowley2018moonshine}
E.~J. Crowley, G.~Gray, and A.~J. Storkey, ``Moonshine: Distilling with cheap
  convolutions,'' in \emph{Advances in Neural Information Processing Systems},
  2018, pp. 2888--2898.

\bibitem{yang2019snapshot}
C.~Yang, L.~Xie, C.~Su, and A.~L. Yuille, ``Snapshot distillation:
  Teacher-student optimization in one generation,'' in \emph{Proceedings of the
  IEEE Conference on Computer Vision and Pattern Recognition}, 2019, pp.
  2859--2868.

\bibitem{li2019learning}
J.~Li, Y.~Wong, Q.~Zhao, and M.~S. Kankanhalli, ``Learning to learn from noisy
  labeled data,'' in \emph{Proceedings of the IEEE Conference on Computer
  Vision and Pattern Recognition}, 2019, pp. 5051--5059.

\bibitem{vahdat2017toward}
A.~Vahdat, ``Toward robustness against label noise in training deep
  discriminative neural networks,'' in \emph{Advances in Neural Information
  Processing Systems}, 2017, pp. 5596--5605.

\bibitem{xiao2015learning}
T.~Xiao, T.~Xia, Y.~Yang, C.~Huang, and X.~Wang, ``Learning from massive noisy
  labeled data for image classification,'' in \emph{Proceedings of the IEEE
  conference on computer vision and pattern recognition}, 2015, pp. 2691--2699.

\bibitem{jiang2017mentornet}
L.~Jiang, Z.~Zhou, T.~Leung, L.-J. Li, and L.~Fei-Fei, ``Mentornet: Learning
  data-driven curriculum for very deep neural networks on corrupted labels,''
  \emph{arXiv preprint arXiv:1712.05055}, 2017.

\bibitem{goldberger2016training}
J.~Goldberger and E.~Ben-Reuven, ``Training deep neural-networks using a noise
  adaptation layer,'' in \emph{ICLR}, 2016.

\bibitem{veit2017learning}
A.~Veit, N.~Alldrin, G.~Chechik, I.~Krasin, A.~Gupta, and S.~Belongie,
  ``Learning from noisy large-scale datasets with minimal supervision,'' in
  \emph{Proceedings of the IEEE Conference on Computer Vision and Pattern
  Recognition}, 2017, pp. 839--847.

\bibitem{sukhbaatar2014training}
S.~Sukhbaatar, J.~Bruna, M.~Paluri, L.~Bourdev, and R.~Fergus, ``Training
  convolutional networks with noisy labels,'' \emph{arXiv preprint
  arXiv:1406.2080}, 2014.

\bibitem{patrini2017making}
G.~Patrini, A.~Rozza, A.~Krishna~Menon, R.~Nock, and L.~Qu, ``Making deep
  neural networks robust to label noise: A loss correction approach,'' in
  \emph{Proceedings of the IEEE Conference on Computer Vision and Pattern
  Recognition}, 2017, pp. 1944--1952.

\bibitem{ren2018learning}
M.~Ren, W.~Zeng, B.~Yang, and R.~Urtasun, ``Learning to reweight examples for
  robust deep learning,'' \emph{arXiv preprint arXiv:1803.09050}, 2018.

\bibitem{lee2018cleannet}
K.-H. Lee, X.~He, L.~Zhang, and L.~Yang, ``Cleannet: Transfer learning for
  scalable image classifier training with label noise,'' in \emph{Proceedings
  of the IEEE Conference on Computer Vision and Pattern Recognition}, 2018, pp.
  5447--5456.

\bibitem{liu2015classification}
T.~Liu and D.~Tao, ``Classification with noisy labels by importance
  reweighting,'' \emph{IEEE Transactions on pattern analysis and machine
  intelligence}, vol.~38, no.~3, pp. 447--461, 2015.

\bibitem{krizhevsky2009learning}
A.~Krizhevsky, ``Learning multiple layers of features from tiny images,'' Tech.
  Rep., 2009.

\bibitem{netzer2011reading}
Y.~Netzer, T.~Wang, A.~Coates, A.~Bissacco, B.~Wu, and A.~Y. Ng, ``Reading
  digits in natural images with unsupervised feature learning,'' in \emph{NIPS
  workshop on deep learning and unsupervised feature learning}, vol. 2011,
  no.~2, 2011, p.~5.

\bibitem{simonyan2014very}
K.~Simonyan and A.~Zisserman, ``Very deep convolutional networks for
  large-scale image recognition,'' in \emph{ICLR}, 2015.

\bibitem{he2016deep}
K.~He, X.~Zhang, S.~Ren, and J.~Sun, ``Deep residual learning for image
  recognition,'' in \emph{CVPR}, 2016.

\bibitem{huang2016densely}
G.~Huang, Z.~Liu, L.~Van Der~Maaten, and K.~Q. Weinberger, ``Densely connected
  convolutional networks,'' in \emph{Proceedings of the IEEE conference on
  computer vision and pattern recognition}, 2017, pp. 4700--4708.

\bibitem{howard2017mobilenets}
A.~G. Howard, M.~Zhu, B.~Chen, D.~Kalenichenko, W.~Wang, T.~Weyand,
  M.~Andreetto, and H.~Adam, ``Mobilenets: Efficient convolutional neural
  networks for mobile vision applications,'' \emph{arXiv preprint
  arXiv:1704.04861}, 2017.

\bibitem{paszke2017automatic}
A.~Paszke, S.~Gross, S.~Chintala, G.~Chanan, E.~Yang, Z.~DeVito, Z.~Lin,
  A.~Desmaison, L.~Antiga, and A.~Lerer, ``Automatic differentiation in
  pytorch,'' 2017.

\bibitem{lee2015deeply}
C.-Y. Lee, S.~Xie, P.~W. Gallagher \emph{et~al.}, ``Deeply-supervised nets.''
  in \emph{AISTATS}, 2015.

\bibitem{romero2015fitnets}
A.~Romero, N.~Ballas, S.~E. Kahou, A.~Chassang, C.~Gatta, and Y.~Bengio,
  ``Fitnets: Hints for thin deep nets,'' in \emph{International Conference on
  Learning Representations}, 2015.

\bibitem{larsson2016fractalnet}
G.~Larsson, M.~Maire, and G.~Shakhnarovich, ``Fractalnet: Ultra-deep neural
  networks without residuals,'' \emph{arXiv preprint arXiv:1605.07648}, 2016.

\bibitem{liu2017learning}
Z.~Liu, J.~Li, Z.~Shen, G.~Huang, S.~Yan, and C.~Zhang, ``Learning efficient
  convolutional networks through network slimming,'' in \emph{Proceedings of
  the IEEE International Conference on Computer Vision}, 2017, pp. 2736--2744.

\bibitem{goodfellow2013maxout}
I.~J. Goodfellow, D.~Warde-Farley, M.~Mirza, A.~Courville, and Y.~Bengio,
  ``Maxout networks,'' in \emph{ICML}, 2013.

\bibitem{krizhevsky2012imagenet}
A.~Krizhevsky, I.~Sutskever, and G.~E. Hinton, ``Imagenet classification with
  deep convolutional neural networks,'' in \emph{Advances in neural information
  processing systems}, 2012, pp. 1097--1105.

\bibitem{ioffe2015batch}
S.~Ioffe and C.~Szegedy, ``Batch normalization: Accelerating deep network
  training by reducing internal covariate shift,'' \emph{arXiv preprint
  arXiv:1502.03167}, 2015.

\bibitem{szegedy2015going}
C.~Szegedy, W.~Liu, Y.~Jia, P.~Sermanet \emph{et~al.}, ``Going deeper with
  convolutions,'' in \emph{CVPR}, 2015.

\bibitem{zhang2018mixup}
H.~Zhang, M.~Cisse, Y.~N. Dauphin, and D.~Lopez-Paz, ``mixup: Beyond empirical
  risk minimization,'' in \emph{International Conference on Learning
  Representations}, 2018.

\bibitem{yun2019cutmix}
S.~Yun, D.~Han, S.~J. Oh, S.~Chun, J.~Choe, and Y.~Yoo, ``Cutmix:
  Regularization strategy to train strong classifiers with localizable
  features,'' \emph{arXiv preprint arXiv:1905.04899}, 2019.

\bibitem{loshchilov2016sgdr}
I.~Loshchilov and F.~Hutter, ``Sgdr: Stochastic gradient descent with warm
  restarts,'' \emph{arXiv preprint arXiv:1608.03983}, 2016.

\end{thebibliography}

\end{document}